\documentclass[10pt,twocolumn,letterpaper]{article}

\usepackage[pagenumbers]{cvpr}

\usepackage{cuted}

\usepackage{currfile}

\usepackage{caption}

\DeclareRobustCommand{\twsname}{\textup{\textsc{Temporal Weight Surgery}}}
\DeclareRobustCommand{\tws}{\textup{TWS}}
\AtBeginDocument{\pdfstringdefDisableCommands{%
  \def\twsname{Temporal Weight Surgery}%
  \def\tws{TWS}%
}}

\usepackage{multirow}
\usepackage{capt-of}
\usepackage{array}
\usepackage{needspace}

\definecolor{cvprblue}{rgb}{0.21,0.49,0.74}
\usepackage[pagebackref,breaklinks,colorlinks,allcolors=cvprblue]{hyperref}

\title{\twsname{}: Repairing Video Grounding with LLM Editors}

\author{%
\begin{tabular}{c@{\hspace{3em}}c}
Ali Asgarov & Chris Thomas\\
{\small\tt aliasgarov@vt.edu} & {\small\tt christhomas@vt.edu}
\end{tabular}\\[0.3em]
{\normalsize Department of Computer Science}\\
{\normalsize Virginia Tech}
}

\providecommand{\nolinenumbers}{}

\begin{document}
\addtocontents{toc}{\protect\setcounter{tocdepth}{-1}}
\twocolumn[{%
\renewcommand\twocolumn[1][]{#1}%
\vspace*{-3.4em}
\maketitle
\vspace{-2.4em}
\begin{center}
\includegraphics[width=\textwidth]{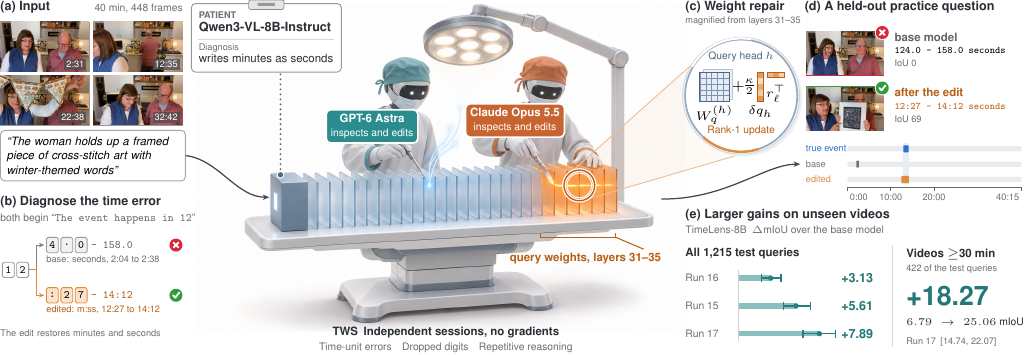}
\captionof{figure}{\textbf{\twsname{} repairs video grounding by diagnosing failures and editing weights without gradients.} Editors work independently. (a--d) In this Qwen3-VL-8B example, Claude traces a time-format error to query weights in layers 31--35. The magnified view shows the rank-1 repair. On a held-out practice query, it moves the answer from two minutes to the event near 13 minutes, raising IoU from 0 to 69 (0--100 scale). Frames and answers are recorded. (e) Three independent Astra sessions repair dropped timestamp digits in TimeLens-8B, gaining 3.13, 5.61 and 7.89 mIoU over the base model on 1,215 VUE-TR and VUE-TR-V2 test queries. Run 17 gains 18.27 on their 422-query subset with videos of at least 30 minutes. Whiskers and brackets give 95\% bootstrap intervals over videos. All sessions and baseline comparisons appear in \cref{tab:editors,sec:results} and the supplement.}

\label{fig:teaser}
\end{center}
\vspace{0.6em}
}]

\begin{abstract}
A surgeon first finds what is wrong, then fixes only that part.
We use this idea for video grounding, where video-language models often find the right moment in a long video but write its time wrong.
In \twsname{} (\tws{}), a general-purpose LLM reads the model's weights and activations, finds where the error arises and repairs it there, without gradients.
Separate sessions find the same cause, and this diagnosis of where and why the model fails is our main contribution.
Three GPT-6 Astra sessions trace a dropped digit in TimeLens-8B to one MLP layer, and the largest edit gains 7.9 mIoU on 1,215 unseen questions and 18.3 on videos of 30 minutes or more.
Two Claude Opus 5.5 sessions trace minutes written as seconds in Qwen3-VL-8B to the separator and gain 2.9 and 1.3.
Before any edit, its answers often have the right minute and the wrong separator, and fixing only the separator lifts their mIoU from 13.1 to 19.2.
Every improving edit beats a random edit of its size, and the edited models run as fast as before.
\end{abstract}

\section{Introduction}
\label{sec:intro}

Temporal video grounding asks a model to return the start and end times of an event described by a text query.
A wrong interval can reflect an error in locating the event or in writing its time.
For an event near 12 minutes, an answer of ``\texttt{12.0 - 13.0 seconds}'' places the interval near the beginning of the video.
The numbers alone do not reveal whether the model looked in the wrong place or expressed minutes as seconds.
These causes call for different repairs.

Video-language models often find the right moment but write the wrong time, especially on long videos.
TimeLens-8B~\cite{zhang2026timelens}, a model trained for grounding, can instead drop a digit, answering 174 to 179 seconds for an event at 1,744 to 1,795 seconds.
We show that a general-purpose LLM can find where in such a model's weights this happens and repair it there, without gradients, and that independent sessions converge on the same cause.
This diagnosis, where in the model the error arises and why, is our main contribution; once it is known, simpler repairs such as standard steering can use it too.
We build on weight-space programming~\cite{takehi2026sparks}, a recent concept in which an LLM reads and edits another network's weights.
In \twsname{} (\tws{}), the editor inspects weights and activations, tests hypotheses on practice questions and submits one fixed weight update, evaluated once on unseen videos of VUE-TR and VUE-TR-V2~\cite{vidi2025,vidi2}.
The editors are Claude Opus 5.5 and GPT-6 Astra (Claude and Astra below).

Why can a few weights repair such an error? Writing a time is a short series of token choices.
After writing the 12 of a start at 12:40, Qwen3-VL-8B must choose the next token, a colon for minutes and seconds or a decimal point for seconds, and its last layers make this choice.
TimeLens-8B's dropped digit is the same kind of choice, a decimal point written one digit too early.
Changing how a few weights respond at that one position changes the written time, and all other weights stay unchanged.

The type of error is visible in the answers themselves, and it suggests simpler corrections, such as reading an early answer in seconds as minutes; some early answers are correct, however, so such a rule can introduce errors.
We therefore test both whether a repair generalizes to unseen videos and whether it improves on standard corrections built on the same diagnosis, which separates finding the failure from building the edit.
Our main findings are as follows.
\begin{enumerate}
  \item \textbf{Editors find where a timestamp error arises and repair it.}
  Three Astra sessions independently trace a dropped digit in TimeLens-8B~\cite{zhang2026timelens} to a decimal point written one digit early and edit the same output direction of one MLP matrix; the largest edit gains $+7.89$ points of mean temporal intersection over union (mIoU) on 1,215 test questions and $+18.27$ on videos of 30 minutes or more.
  Two Claude sessions trace minutes written as seconds in Qwen3-VL-8B-Instruct~\cite{bai2025qwen3vl} to the choice between a colon and a decimal point in its last five layers, and gain $+2.88$ and $+1.30$ by editing layer 35's value and output projections (the value/output edit) or five query projections (the query edit).
  An editor shown only the answers finds the same errors but not where they arise, and its rules gain less.
  \item \textbf{Qwen3-VL-8B holds more of the time than it writes.}
  Before any edit, the first number it writes is the true start minute in 45\% of the answers with a close separator choice, against 16\% by chance, and fixing only the separator raises their mIoU from 13.1 to 19.2, against 9.8 for reading the written times as minutes (\cref{sec:separator}).
  \item \textbf{The repairs work.}
  Every edit that improves on the base model beats a random edit of its size, and the best edit beats steering built on the same diagnosis on TimeLens-8B and Qwen3-VL-2B, by $+3.03$ and $+1.24$; on Qwen3-VL-8B, once the failure is located, standard steering is level with the edits (\cref{fig:test-comparison}).
  The edits add no inference computation.
\end{enumerate}

\section{Related work}
\label{sec:related}

\paragraph{Temporal video grounding.}
TimeLens~\cite{zhang2026timelens} and TimeLens2~\cite{zhu2026timelens2} train Qwen-VL models to return time spans directly.
General video-language models such as Qwen3-VL~\cite{bai2025qwen3vl} can also answer grounding queries without being specialised for the task.
VUE-TR and VUE-TR-V2~\cite{vidi2025,vidi2} evaluate this ability on videos ranging from under a minute to over an hour.
Du~\etal~\cite{du2026know} find attention heads that locate the queried interval even when the written answer misses it, and at inference, when a confidence gate judges the first answer unreliable, restrict the video to the interval these heads attend to and let the model answer again; we find errors in how the time is written and repair them in the weights.
We trace errors in these answers to specific computations and repair them with direct weight edits.

\paragraph{Model editing and activation steering.}
ROME~\cite{meng2022rome} and MEMIT~\cite{meng2023memit} write rank-1 or batched updates to factual associations in MLP layers.
MEND~\cite{mitchell2022mend} trains small networks to transform gradients into edits.
Task arithmetic~\cite{ilharco2023task} adds or subtracts task vectors, the differences between fine-tuned and pre-trained weights, while Text-to-LoRA~\cite{charakorn2025t2l} generates adapters from task descriptions.
Activation steering changes hidden representations along selected directions~\cite{turner2023steering}.
Weight orthogonalization implements directional ablation directly in the parameters~\cite{arditi2024refusal}.
Automated interpretability addresses the diagnosis itself.
MAIA~\cite{shaham2024maia} uses a multimodal LLM to design image experiments that explain components of vision models.

\paragraph{An LLM as the weight editor.}
Weight-space programming~\cite{takehi2026sparks} treats neural weights as programs an LLM can inspect and modify.
The original experiments prohibit gradients, as do ours.
Hu~\etal~\cite{hu2026exploring} let GPT-6 Astra edit Qwen3.5-0.8B and Qwen3.5-9B on seven benchmarks.
Zhang~\etal~\cite{zhang2026debug} use GPT-6 Astra with simulator feedback to edit token-embedding rows of the $\pi_{0.5}$ robot policy.
For smaller networks, Nusrat and Nusrat~\cite{nusrat2026clinical} use Claude Opus 5.5 to audit and repair clinical decision models with 433 to 4,801 parameters, and Lindholz~\etal~\cite{lindholz2026medical} have Claude Opus 5.5 and GPT-6 Astra write every weight of a small medical-imaging CNN of fixed design without training data.
Ahmed~\cite{ahmed2026reason} questions what weight inspection reveals, reporting that Claude Opus 5.5 interprets small CNNs through local arithmetic and memorised dataset priors.
In \tws{}, both editors locate and repair failures in multi-billion-parameter video-language models, and explicit controls separate finding a failure from designing its repair.

\section{Setting and evaluation protocol}
\label{sec:protocol}

\begin{figure*}[t]
\centering
\includegraphics[width=\linewidth]{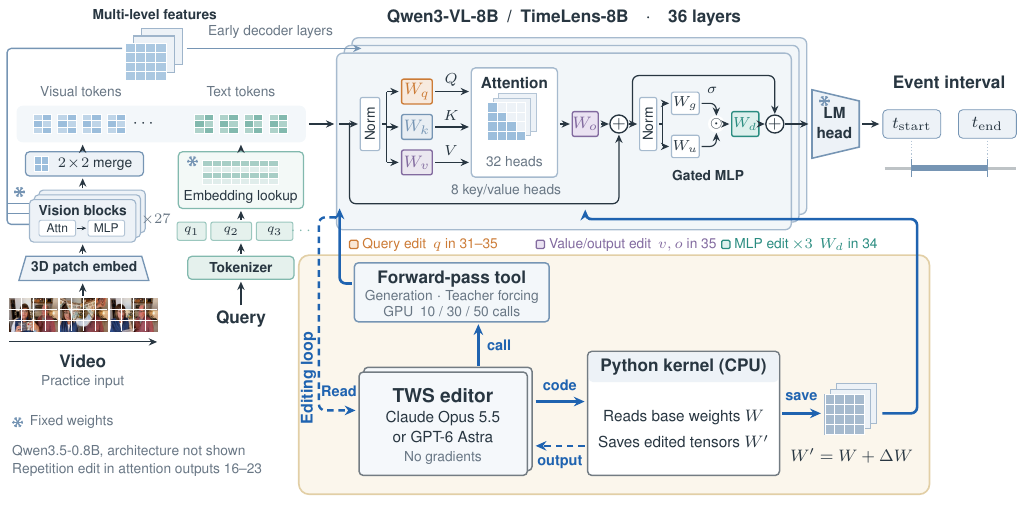}
\caption{\textbf{The \tws{} editor diagnoses failures and writes weight updates through two tools.} In the shaded editing loop, the editor calls the GPU forward-pass tool, which runs practice batches by generation or teacher forcing, reads the scores and activations, writes code in the CPU Python kernel and saves complete edited tensors $W'$, which the next call loads. The shown architecture is shared by Qwen3-VL-8B and TimeLens-8B. Orange and purple mark the independent Qwen query and value/output edits. Teal marks the three TimeLens MLP edits. The bottom note marks Qwen3.5-0.8B's repetition edit. The 1,215 test questions stay hidden from the editor. Supplementary Fig.~S1 gives diagnostic details.}
\label{fig:pipeline}
\end{figure*}

The editor is a general-purpose LLM that changes a video-language model's weights through two tools, without gradients, and never sees the test sets (\cref{fig:pipeline}).

\subsection{Task, data and test}

\paragraph{Models.}
The two main models are TimeLens-8B~\cite{zhang2026timelens}, a grounding model fine-tuned from Qwen3-VL-8B-Instruct, and Qwen3-VL-8B-Instruct itself~\cite{bai2025qwen3vl}; both have a 36-layer language model of width 4,096 with 32 query and 8 key/value heads.
Each receives 448 evenly spaced frames and a prompt requesting ``\texttt{The event happens in \textless start time\textgreater{} - \textless end time\textgreater{} seconds}'', and answers greedily with at most 256 new tokens.
We measure mIoU, the mean temporal intersection over union between predicted and true intervals, on a 0--100 scale.

\paragraph{Practice data.}
The editor can use 480 labelled questions on 227 videos of at least 10 minutes from the public TimeLens2-93K training set~\cite{zhu2026timelens2}, none of them in the test sets, on which Qwen3-VL-8B scores 21.68 mIoU.
Sessions hold out part of these questions to check their edits, and we call the rest development questions.

\paragraph{Unseen test videos.}
Each submitted edit is evaluated once, under a rule fixed before the session, on the 1,215 visual questions of VUE-TR and VUE-TR-V2~\cite{vidi2025,vidi2} whose videos we could download, from under a minute to over an hour long (supplementary Sec.~A).
The edited model and the base model, the model before the edit, are evaluated in the same job, and the primary outcome is their paired mIoU difference, pooled over both sets.
One fixed parser, that of TimeLens~\cite{zhang2026timelens}, reads every method's answers. It reads m:ss and h:mm:ss times as minutes and seconds whatever unit follows, other numbers as seconds, and an answer without a valid span scores zero.
We report improvement or degradation when the 95\% paired bootstrap interval over videos lies above or below zero; an interval containing zero is denoted \emph{not significant} (NS) and does not establish equivalence.

\subsection{What the editor can see and change}

A session is one conversation with the editor, which ends with at most one submitted edit.
In a typical session the editor runs the model on practice questions, reads the answers, log-probabilities and hidden states, states a hypothesis, writes Python that builds a candidate edit from the stored weights and tests it with the next call, before it submits one edit.
The editor, Claude Opus 5.5 at maximum reasoning effort or GPT-6 Astra, has two tools: a Python kernel that reads the model files and writes edited tensors, and a forward-pass tool that runs the model on practice questions and returns paired scores, log-probabilities or hidden states.
In the attention-only setting it may change the $q$, $k$, $v$ and $o$ projections of all 36 layers with 10 forward-pass calls; in the full-weight setting, every tensor except the embeddings, output layer and final norm with 50 calls (\cref{tab:editors}).
Gradients, backward passes, optimisers and iterative fitting to a loss are prohibited; closed-form fits on recorded activations are allowed, and every network measurement must go through the forward-pass tool.
Supplementary Sec.~A gives the call modes, limits and test details, and how the setting follows the weight-space programming report~\cite{takehi2026sparks}.

\section{Diagnosing and repairing timestamp errors}
\label{sec:method}

\tws{} editors identify two distinct failures, a dropped digit in TimeLens-8B and minutes written as seconds in Qwen3-VL-8B.
We first explain the three Astra sessions on TimeLens-8B, then the two independent Claude sessions on Qwen3-VL-8B.
All measurements in this section use practice questions rather than the test sets.

\subsection{MLP editing for a dropped digit}

On long videos, TimeLens-8B can drop a digit from its timestamps, answering 174 to 179 seconds for an event at 1,744 to 1,795 seconds.
Three Astra sessions in the full-weight setting find this failure on their own and trace it to a decimal point written one digit too early.
All three submit a rank-1 edit to layer 34's MLP output projection, a change along one input and one output direction of the matrix.
The output direction is the output layer's direction for ``\texttt{.}'' relative to the digits, and the edit lowers the MLP's output along it.
The three output directions align ($|\cos| \geq 0.99$), while their input directions, which decide when the edit acts, and their sizes differ.
The largest is $W' = W - 8\,u\,(v^{\top} W)$ for that direction $u$ and a unit selector $v$ of residual states at premature decimal points, made orthogonal to the states at correct decimal points and at the answer's other tokens (supplementary Sec.~B).
In plain terms, the edit makes a decimal point less likely when the model's state looks like one before a premature decimal point, and leaves correct decimal points and other tokens alone.
On held-out practice questions the three edits gain $+10.72$, $+7.32$ and $+6.30$ mIoU (\cref{tab:editors}), and one gains only $+1.26$ when every frame is black, so the repair still depends on the video.

\subsection{Query editing for minutes written as seconds}

\paragraph{Minutes are written as seconds.}
Of the base model's 480 practice answers, 131 end within the first 2\% of the video and 83 of these have zero IoU.
For many such answers, reading the numbers as minutes moves them towards the true event.
Answers already written in m:ss (minutes and seconds) have rank correlation 0.80 to 0.85 with the true times, against 0.23 to 0.25 for seconds answers.

\paragraph{The final layers choose the separator.}
The editor traces the failure to the separator, the token after the start time's first number, where the model writes ``\texttt{.}'' instead of ``\texttt{:}''.
In answers with minutes written as seconds, ``\texttt{.}'' beats ``\texttt{:}'' by a logit margin of about 6, against 19 to 25 in genuine seconds answers.
A logit lens reads each layer's residual stream through the final norm and output layer~\cite{nostalgebraist2020logitlens,belrose2023tuned}.
It places the separator choice in layers 31 to 35.
A linear probe from layer 20 onwards reads m:ss in 81\% of these answers, so the model encodes the format before its final layers write seconds, and \cref{sec:separator} tests whether it also holds the time.

\paragraph{A query edit guided by a format probe.}
For each head $h$ in layers 31 to 35, let $\delta q_h$ be the difference between its queries at separator positions associated with m:ss and seconds representations.
Let $r_\ell$ be a ridge-regression probe of the normalised residual stream entering layer $\ell$, with target $+1$ for m:ss, $-1$ for seconds and approximately 0 at other positions.
The editor adds a rank-1 term to each head's query-projection rows
\begin{equation}
  W_q^{(h)} \leftarrow W_q^{(h)} + \tfrac{\kappa}{2}\, \delta q_h\, r_\ell^{\top} .
\end{equation}
The probe sets how far each query moves towards its m:ss direction, so positions with little format signal change little.
The submitted strength is $\kappa = 4$, changing the five query-projection tensors by 31\% to 34\% of their norms.

\subsection{Value/output editing for the same error}

A second Claude session reaches the same diagnosis, finding that decimal answers after about 5 minutes are minutes written as seconds.
It targets the same separator decision but changes layer 35's value and output projections instead of its queries (the value/output edit).
In each of the 8 key/value groups, two value directions read residual dimensions that are almost constant on text tokens, giving every head nearly constant channels, and a ridge fit sets the output weights of all 32 heads so that these channels add a common steering vector to the residual stream, whatever the attention pattern.
At a typical decision position, this vector raises the ``\texttt{:}'' logit by about 6 while leaving the direct logits of ``\texttt{.}'', digits and other answer tokens unchanged.

Of 1, 1.5 and 2 times the initial strength, 1.5 scores best on all 480 practice questions (25.98 mIoU against 21.68); at 2 the model replaces correct decimal points.
The submitted edit changes the two tensors by 5.5\% and 13\% of their norms.
Removing the steering component from this edit gives $-0.36 \pm 0.35$ on all 480 questions, attributing the gain to that component.

\section{Results}
\label{sec:results}

All test comparisons use the criterion in \cref{sec:protocol}.

\begin{table*}[t]
\centering
\caption{\textbf{Every \tws{} edit (ours) that improves on the base model beats a random edit of its size.} Calls count batched requests to the forward-pass tool (10 for attention weights, 30 or 50 for all weights). Parentheses give MLP Frobenius norms relative to the base tensors. Practice $\Delta$mIoU uses the session's subset and one standard error; $\dagger$ marks reuse for strength selection, $\ddagger$ thinking mode. Test gains pool VUE-TR and VUE-TR-V2, and the last column compares each edit with a random edit of its size, both with 95\% bootstrap intervals over videos; dashes mean unevaluated. Run numbers follow supplementary Table~S3, which lists all 21 sessions. Bold marks each model's best.}
\label{tab:editors}
\footnotesize
\setlength{\tabcolsep}{1.7pt}
\begin{tabular}{rlllllll}
\toprule
Run & Editor & Base model & Calls & Weight update & \shortstack{Practice\\$\Delta$mIoU} & \shortstack{Test\\$\Delta$mIoU} & \shortstack{Edit vs. control\\$\Delta$mIoU} \\
\midrule
17 & GPT-6 Astra & TimeLens-8B & 50 & rank-1 MLP output, layer 34 (0.15) & $\boldsymbol{+10.72} \pm 1.60$ & $\boldsymbol{+7.89}$ $[+6.28, +9.46]$ & $\boldsymbol{+8.17}$ $[+6.56, +9.79]$ \\
15 & GPT-6 Astra & TimeLens-8B & 50 & rank-1 MLP output, layer 34 (0.09) & $+7.32 \pm 1.27$ & $+5.61$ $[+4.35, +7.02]$ & $+5.49$ $[+4.22, +6.87]$ \\
16 & GPT-6 Astra & TimeLens-8B & 50 & rank-1 MLP output, layer 34 (0.04) & $+6.30 \pm 1.71$ & $+3.13$ $[+2.19, +4.14]$ & $+3.44$ $[+2.51, +4.47]$ \\
4 & Claude Opus 5.5 & Qwen3-VL-8B & 10 & value/output edit, layer 35 & $\boldsymbol{+3.96} \pm 1.37^{\dagger}$ & $\boldsymbol{+2.88}$ $[+1.85, +3.97]$ & $\boldsymbol{+2.77}$ $[+1.75, +3.84]$ \\
3 & Claude Opus 5.5 & Qwen3-VL-8B & 10 & query edit, layers 31--35 & $+3.77 \pm 1.11$ & $+1.30$ $[+0.63, +2.01]$ & $+1.95$ $[+1.10, +2.87]$ \\
12 & GPT-6 Astra & Qwen3-VL-2B & 50 & two head outputs, layers 16, 20 & $\boldsymbol{+4.40} \pm 1.27$ & $\boldsymbol{+2.11}$ $[+1.36, +2.89]$ & $+2.57$ $[+1.83, +3.34]$ \\
14 & GPT-6 Astra & Qwen3-VL-2B & 50 & two head outputs, layer 19 & $+1.02 \pm 1.08$ & $+1.49$ $[+0.63, +2.40]$ & $\boldsymbol{+2.97}$ $[+1.97, +4.01]$ \\
9 & GPT-6 Astra & Qwen3-VL-2B & 50 & MLP output, layer 25 & $+1.31 \pm 1.27$ & $-0.12$ $[-1.04, +0.82]$ & -- \\
18 & GPT-6 Astra & Qwen3.5-0.8B$^{\ddagger}$ & 30 & rank-1 attention outputs, layers 16--23 & $\boldsymbol{+3.95} \pm 1.79$ & $\boldsymbol{+2.33}$ $[+0.88, +3.79]$ & $\boldsymbol{+2.06}$ $[+0.52, +3.63]$ \\
19 & GPT-6 Astra & Qwen3.5-0.8B$^{\ddagger}$ & 30 & head 7 of $o$ halved, layer 15 & $-0.85 \pm 1.59$ & $+1.24$ $[-0.14, +2.70]$ & -- \\
\bottomrule
\end{tabular}
\end{table*}

\begin{figure}[t]
\centering
\includegraphics[width=\linewidth]{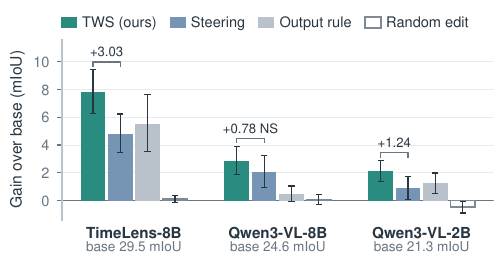}
\caption{\textbf{\tws{} beats a random edit of its size on every model, and steering on two of three.} Test gains over the base model on the 1,215 test questions with 95\% bootstrap intervals over videos, all on one machine against the same base answers (run 17 gains $+7.82$ here, $+7.89$ in \cref{tab:editors}). Brackets give \tws{} minus steering. On Qwen3-VL-2B the output rule is level with \tws{} ($+0.88$, NS).}
\label{fig:test-comparison}
\end{figure}

\paragraph{Baselines.}
Each model's best edit is compared with three alternatives (\cref{fig:test-comparison}), and \cref{tab:main} gives every method's test mIoU, where \tws{} is highest on all three models.
The output rule applies the diagnosis to the answers. For Qwen, a plain-seconds answer ending in the first 2\% of a video of 10 minutes or more is read as minutes, and for TimeLens-8B the times-ten rule multiplies both times by ten when the answer ends in the first tenth of the video.
A random edit of the edit's size tests whether the edit's location and direction matter, and steering, built on the same diagnosis without an LLM, tests whether the LLM's own edit adds to the diagnosis.
Steering applies activation steering~\cite{turner2023steering} with a difference-of-means direction~\cite{arditi2024refusal} as a weight edit, $W' = W + \alpha\, u_l u_l^{\top} W$ in the attention and MLP outputs of five consecutive layers, where $u_l$ is the normalised difference between mean hidden states before the right and the wrong token (m:ss against decimal seconds, or a fourth integer digit against a decimal point), with the layers and $\alpha$ chosen on development questions (supplementary Table~S1).

\begin{table}[t]
\centering
\caption{\textbf{\tws{} gives the highest test mIoU on all three models.} Test mIoU on VUE-TR (TR), VUE-TR-V2 (V2) and both pooled (All, 1,215 questions), from the answers of \cref{fig:test-comparison}, all on one machine. \tws{} is each model's best edit (runs 17, 4 and 12), and the random edit has its size. $^{\ast}$The times-ten rule for TimeLens-8B. Best in bold.}
\label{tab:main}
\footnotesize
\setlength{\tabcolsep}{2.1pt}
\begin{tabular}{lccccccccc}
\toprule
 & \multicolumn{3}{c}{TimeLens-8B} & \multicolumn{3}{c}{Qwen3-VL-8B} & \multicolumn{3}{c}{Qwen3-VL-2B} \\
\cmidrule(lr){2-4}\cmidrule(lr){5-7}\cmidrule(lr){8-10}
Method & TR & V2 & All & TR & V2 & All & TR & V2 & All \\
\midrule
Base model & 34.6 & 25.9 & 29.5 & 29.8 & 20.9 & 24.6 & 25.3 & 18.4 & 21.3 \\
Output rule$^{\ast}$ & 40.4 & 31.2 & 35.1 & 30.4 & 21.3 & 25.1 & \textbf{27.7} & 18.8 & 22.5 \\
Random edit & 34.5 & 26.1 & 29.7 & 29.7 & 21.1 & 24.7 & 24.9 & 17.9 & 20.8 \\
Steering, no LLM & 38.8 & 31.1 & 34.3 & 31.5 & \textbf{23.2} & 26.7 & 27.2 & 18.5 & 22.2 \\
\midrule
\tws{} (ours) & \textbf{42.7} & \textbf{33.5} & \textbf{37.4} & \textbf{33.3} & \textbf{23.2} & \textbf{27.5} & 27.5 & \textbf{20.4} & \textbf{23.4} \\
\bottomrule
\end{tabular}
\end{table}

\subsection{Three TimeLens-8B sessions find one repair}
\label{sec:timelens}

\begin{figure}[t]
\centering
\includegraphics[width=\linewidth]{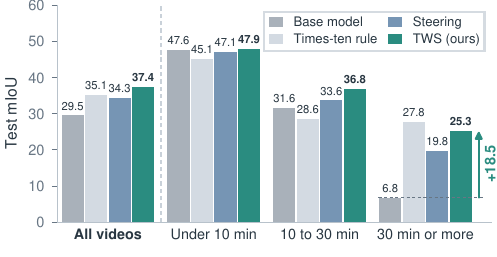}
\caption{\textbf{\tws{} gains most over all TimeLens-8B test videos, and repairs long ones without harming short ones.} Test mIoU overall and by video length, from the answers of \cref{fig:test-comparison}. The arrow marks the gain of \tws{} (run 17) on videos of 30 minutes or more, $+18.51$ $[+15.06, +22.24]$; under 10 minutes it is level with the base model ($+0.30$, NS). The times-ten rule gains more on the longest videos but loses on shorter ones, where it also stretches correct early answers.}
\label{fig:timelens-length}
\end{figure}

TimeLens-8B~\cite{zhang2026timelens} scores only 6.79 mIoU on test videos of 30 minutes or more, often dropping a digit.
The three Astra edits of \cref{sec:method}, runs 17, 15 and 16, gain $+7.89$, $+5.61$ and $+3.13$ across the test sets (\cref{tab:editors}).
On videos of 30 minutes or more, the largest edit raises mIoU from 6.79 to 25.06, a gain of $+18.27$ $[+14.74, +22.07]$ (\cref{fig:timelens-length}; supplementary Fig.~S2).
There the base model finds almost nothing in the last fifth of a video (0.0 mIoU, 23.5 with the largest edit), and overall this edit reaches about two thirds of the $+12.0$ a perfect repair could give (supplementary Sec.~D).
Each beats a random edit of its size (\cref{tab:editors}), and the largest also beats three more random edits. On Charades-TimeLens and MomentSeeker the edits are level with the base model.

The times-ten rule gains $+5.55$ overall.
The largest edit exceeds it by $+2.34$ $[+0.81, +3.81]$, the middle edit shows no significant difference, and the smallest performs worse.
On videos of 30 minutes or more, however, the largest edit does not differ significantly from the rule, so its advantage comes from shorter videos, where the rule also stretches correct early answers (\cref{fig:timelens-length}).
Steering gains $+4.79$ $[+3.45, +6.23]$; the largest edit adds $+3.03$ $[+1.78, +4.31]$ over it, the middle edit is level with it and the smallest falls below it.
A prompt asking for the time in seconds gains only $+1.04$ here, and every edit exceeds it.
At the early decimal point, a decoding rule that writes the model's best digit instead of a decimal point winning by less than 10 gains $+6.35$, level with the times-ten rule, and the largest edit exceeds it by $+1.47$ $[+0.61, +2.38]$; the added digit is rarely the true one (6 of 428 starts), so TimeLens-8B holds the size of the time but not the dropped digit.

\subsection{Two Qwen3-VL-8B sessions, one diagnosis}
\label{sec:qwen}

\begin{figure}[t]
\centering
\includegraphics[width=\linewidth]{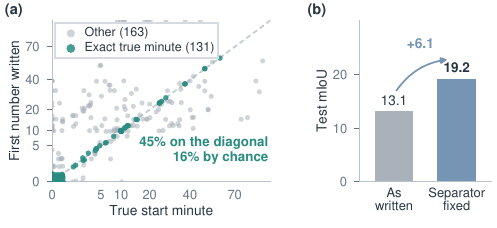}
\caption{\textbf{Before any edit, Qwen3-VL-8B often writes the right minute with the wrong separator.} (a) The first number the base model writes against the true start minute (square-root axes), for the 294 test answers whose ``\texttt{:}'' logit lies within 15 of the top token at the separator. It is the true minute in 131 (45\%, against 16\% by chance), 95 in the first minute and 36 later. (b) Their mIoU as written and with ``\texttt{:}'' forced and the rest written by the model, no weight changed; for an event at 15:34 to 15:40, ``\texttt{15.0 - 18.0 seconds}'' (IoU 0) becomes ``\texttt{15:35 - 15:40 seconds}'' (IoU 83).}
\label{fig:separator}
\end{figure}

\begin{figure*}[t]
\centering
\includegraphics[width=\linewidth]{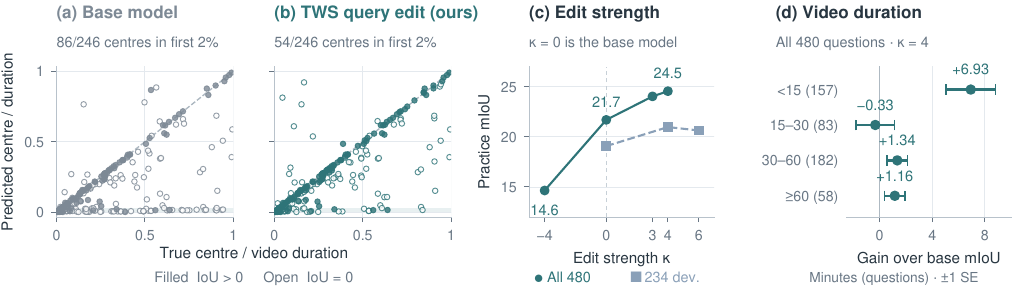}
\caption{\textbf{How the \tws{} query edit changes answers and accuracy.} (a, b) Predicted against true event centres, as shares of the video, on 246 held-out practice questions; the diagonal marks correct centres, shading the first 2\%, where predictions fall from 86 to 54, and open markers zero IoU. (c) Practice mIoU by edit strength; $\kappa=-4$ reverses the edit. (d) Paired gains at $\kappa=4$ by video duration (questions in parentheses, one standard error).}
\label{fig:mechanism}
\end{figure*}

\begin{figure*}[t]
\centering
\includegraphics[width=\linewidth]{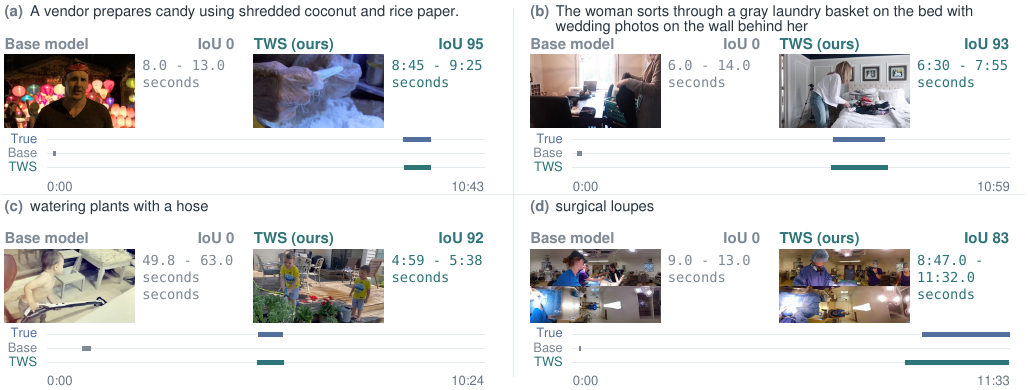}
\caption{\textbf{Timestamp repair moves the answer to the event.} Each example shows the query, frames at the centres of the base and edited answers, and the intervals (blue true, grey base, teal \tws{} query edit). Answers keep the model's formatting; the parser reads 8:45 as minutes and seconds whatever unit follows. These are the four largest IoU gains among held-out practice answers that changed from seconds to m:ss.}
\label{fig:qualitative}
\end{figure*}

For the query edit, the forward-pass tool pairs each edited answer with the base model's answer to the same question (\cref{fig:mechanism}; supplementary Table~S4).
On the 246 held-out practice questions, on videos not used for fitting, the gain is $+3.77 \pm 1.11$ mIoU.
Across all 480 questions, 74 answers switch to m:ss and 3 become incorrect.
Among the 240 questions on videos of 30 minutes or more, answers ending in the first 2\% of the video fall from 98 to 58.
The largest gain occurs below 15 minutes, $+6.93 \pm 1.89$.
Reversing the edit lowers mIoU by $7.04 \pm 0.98$ and changes 123 answers from m:ss to seconds (held-out examples in \cref{fig:qualitative}).

On the test questions, the query edit gains $+1.30$ $[+0.63, +2.01]$.
The output rule shows no significant gain over the base model, $+0.49$ $[-0.07, +1.04]$, and the query edit does not significantly outperform it, $+0.81$ $[-0.02, +1.66]$.
Prompts asking for seconds or for m:ss, the form the edits favour, lower mIoU by $1.86$ and $3.77$ (supplementary Table~S2).

The second Claude session, run 4, starts from scratch with the same task, questions and budget, and finds the same error (\cref{sec:method}).
The test gain is $+2.88$ $[+1.85, +3.97]$.
It exceeds the query edit by $+1.58$ $[+0.55, +2.59]$ and the output rule by $+2.39$ $[+1.29, +3.52]$.

On a second machine, where comparisons stay within the machine since bfloat16 greedy answers differ slightly, the two edits gain $+1.91$ and $+2.85$ (\cref{tab:main}).
Four random edits there show no significant gain ($-2.66$ to $+0.08$).
Every edit in \cref{tab:editors} that improves on the base model also beats a random edit of its size (last column).
Steering gains $+2.07$ $[+0.92, +3.27]$, and neither Qwen3-VL-8B edit differs significantly from it ($-0.16$ $[-1.26, +0.92]$ and $+0.78$ $[-0.25, +1.84]$), so once the failure is located, standard steering can repair it on this model.

\subsection{Qwen3-VL-8B holds more of the time than it writes}
\label{sec:separator}

The query-edit session's hypothesis is that ``\emph{the plan representation knows these are m:ss times and only the last step drops the colon}''.
We test it without changing any weight. In the base Qwen3-VL-8B, we set the separator to ``\texttt{:}'' wherever the ``\texttt{:}'' logit lies within 15 of the top one, a margin chosen on practice questions, and let the model write the rest.
This changes 294 test answers.
In 45\% of them the number written before the separator is already the true start minute, against 16\% by chance, and with the separator fixed the model writes times that score 19.2 mIoU, against 13.1 as written and 9.8 when the written times are read as minutes (\cref{fig:separator}; supplementary Sec.~D).
On these answers the model holds more of the time than it writes, and correcting one token recovers part of it.
As a decoding rule this gains $+1.47$ $[+0.65, +2.33]$ over all test questions, level with the query edit and steering, while the value/output edit exceeds it by $+1.38$ $[+0.44, +2.36]$.
TimeLens-8B, by contrast, does not hold its dropped digit (\cref{sec:timelens}).

\subsection{Cross-dataset generalization}

On MomentSeeker~\cite{yuan2025momentseeker} (279 questions on videos of 10 minutes or more), neither Qwen edit differs significantly from the base model.
On 297 short Charades-TimeLens questions~\cite{gao2017tall,zhang2026timelens}, the query edit shows no significant change and the value/output edit degrades mIoU, $-1.01$ $[-1.58, -0.51]$ (supplementary Fig.~S2).
The repaired errors are rare there: these Charades-TimeLens clips last under a minute, and on MomentSeeker the edits raise Qwen3-VL-8B's m:ss answers from 152 to 204 and 239 of 279, but the events are seldom found (supplementary Sec.~D).

\subsection{Editing outcomes across models}

GPT-6 Astra runs with the weight-space programming report's configuration and one-line instruction.
Of all 21 sessions (supplementary Table~S3), each submitting one edit evaluated once, 8 improve the test sets, 8 show no significant change, 1 degrades them and 4 end without a tested edit; none of Astra's five sessions on Qwen3-VL-8B yields a test gain.
Picking each model's session with the highest practice gain, which uses no test result, selects its best test edit on all four models with several sessions.
A screen on 120 practice questions starts a session only if a repair could gain at least 5 mIoU; Astra sessions on two models it turns down, InternVL3.5-8B and MiMo-VL-7B, gain nothing ($+0.11$, NS) or submit no edit (supplementary Sec.~B).
We will release the harness, every session's records, the edits and the scoring scripts.
Of three Qwen3-VL-2B sessions, two gain $+2.11$ and $+1.49$ and beat random edits of their size.
Steering built the same way gains $+0.87$ $[+0.08, +1.74]$ on Qwen3-VL-2B; the better edit adds $+1.24$ $[+0.21, +2.27]$ over it, the other is level with it.
In thinking mode, where 37\% of Qwen3.5-0.8B's~\cite{qwen2026qwen35} answers loop, Astra removes a repetition direction from eight attention outputs, and gains $+2.33$, level with budget forcing~\cite{muennighoff2025s1} (supplementary Fig.~S3).

\section{Discussion}
\label{sec:discussion}

\paragraph{What \tws{} finds beyond the answers.}
The answers alone reveal both errors. Shown only practice answers and true times, Claude finds them, but its rules gain $+6.04$ on TimeLens-8B and $+0.56$ (NS) on Qwen3-VL-8B, below the largest TimeLens edit by 1.79 and the Qwen edits by 2.29 and 1.35 (supplementary Sec.~A), and asking for the format repairs neither error.
The sessions add what the answers do not show. They find where the error arises and can be repaired, at a decimal point written one digit early in TimeLens-8B's layer-34 MLP and at the separator in Qwen3-VL-8B's last five layers. They show that Qwen3-VL-8B holds more of the time than it writes (\cref{sec:separator}). They also build an edit of one to five tensors at that point, whose direction the three TimeLens sessions find independently.
The best Astra edits on TimeLens-8B and Qwen3-VL-2B exceed steering built from that diagnosis, while the Claude edits on Qwen3-VL-8B do not differ significantly from it, so how the edit is built can also matter.

\paragraph{Where errors remain.}
Most remaining errors are not writing errors: 40 to 47\% of test answers miss the event entirely (supplementary Sec.~D).
On videos of 30 minutes or more, the heads that Du~\etal\ find to locate the queried interval~\cite{du2026know} still peak inside 42 to 56\% of the missed events, but strengthening them through the weights lowers accuracy.

\begin{table}[t]
\centering
\caption{\textbf{What each repair costs and gains.} GPU hours of a session's forward-pass calls (L40S for Claude, B200 or A40 for Astra, A40 for steering; wall-clock hours in parentheses; the editor's own LLM calls excluded) or of LoRA's training on one L40S. Gains are paired with the base model on each machine.}
\label{tab:cost}
\footnotesize
\setlength{\tabcolsep}{2pt}
\begin{tabular}{lccc}
\toprule
Repair & GPU h (session h) & Tensors & Test $\Delta$mIoU \\
\midrule
\multicolumn{4}{l}{\emph{Qwen3-VL-8B}} \\
\tws{} (ours), Claude & 3.7, 4.3 (11.5, 14.0) & 5, 2 & $+1.30$, $+2.88$ \\
Steering, no LLM & 4.2 & 10 & $+2.07$ \\
LoRA, same questions & 2.9 & 144 & $+12.97$ \\
Format prompt & none & 0 & $-1.86$ \\
\midrule
\multicolumn{4}{l}{\emph{TimeLens-8B}} \\
\tws{} (ours), Astra & 2.3--21.8 (5.0--25.6) & 1 & $+3.13$--$+7.89$ \\
Steering, no LLM & 6.3 & 10 & $+4.79$ \\
Times-ten rule & none & 0 & $+5.55$ \\
Format prompt & none & 0 & $+1.04$ \\
\bottomrule
\end{tabular}
\end{table}

\paragraph{Editing and supervised fine-tuning.}
Supervised fine-tuning with rank-16 LoRA~\cite{hu2022lora} on the same 480 practice questions gains $+12.97$ mIoU on Qwen3-VL-8B (\cref{tab:cost}), far more than the direct edits, for GPU time of the same order as a session's forward-pass calls, but it changes all 144 attention tensors and gives no account of the failure.
Direct editing instead offers an explanation tied to an edit of one to five tensors, made without gradients.

\paragraph{Limits of the evidence.}
With few sessions per setting, these are session comparisons rather than a ranking of editors.
All practice videos last at least 10 minutes. An edit's site need not be where the error first arises, and few tensors need not mean a small change (the query edit moves its five by 31 to 34\%).
On 200 text questions about decimals and times, the Qwen edits and the largest TimeLens edit match their base models, but the smallest TimeLens edit drops decimal points (56.5\% against 80.0\% correct; supplementary Sec.~A); general video question answering is untested.

\section{Conclusion}
\label{sec:conclusion}

We presented \tws{}, in which a general-purpose LLM reads a video-language model's weights and activations, tests possible causes on practice questions and edits a few weight tensors without gradients.
Three GPT-6 Astra sessions repair a dropped digit in TimeLens-8B along the same weight direction and gain up to $+7.89$ mIoU, and $+18.27$ on videos of 30 minutes or more.
Two Claude sessions trace minutes written as seconds in Qwen3-VL-8B to its last five layers and gain $+2.88$ and $+1.30$, and before any edit this model often writes the right minute with the wrong separator.
An editor that sees only the answers finds the same errors but not where they arise or why.
Once the failure is located, standard steering also repairs Qwen3-VL-8B, and the best Astra edits add $3.03$ and $1.24$ points over it on TimeLens-8B and Qwen3-VL-2B.
Because each repair names its tensors and the decision it changes, it can be tested against a random edit of its size and reused by simpler repairs such as steering.
Other output decisions of video models, such as units, formats and counts, could be diagnosed and repaired in the same way, using the same random-edit and steering controls.
\tws{} shows that a general-purpose LLM, without gradients, can find where in its weights a video-language model writes the time wrong, and repair it there in one to five weight tensors, at no inference cost.

\clearpage
\section*{LLM Usage Statement}
Claude Opus 5.5 and GPT-6 Astra are the weight editors studied in this paper and their edits, explanations and quoted reasoning are experimental outputs (Secs.~\ref{sec:protocol}--\ref{sec:results}), evaluated with tests the authors designed and fixed in advance.
Separately, we used Claude (Anthropic) and ChatGPT (OpenAI) to generate parts of the code, figures and writing, which the authors reviewed, edited and approved.
GPT-6 Astra's image generation rendered the illustration in \cref{fig:teaser}.
For the experiments and for research ideas, hypotheses, claims and conclusions, the tools only assisted by editing or refining content produced by the authors.
No tool was used for data creation or annotation.
The authors take full responsibility for all content.

\clearpage
\addtocontents{toc}{\protect\setcounter{tocdepth}{2}}
\renewcommand{\topfraction}{.9}\renewcommand{\bottomfraction}{.6}\renewcommand{\textfraction}{.08}
\renewcommand{\floatpagefraction}{.8}\renewcommand{\dbltopfraction}{.9}\renewcommand{\dblfloatpagefraction}{.8}
\setcounter{topnumber}{1}\setcounter{bottomnumber}{1}\setcounter{totalnumber}{2}\setcounter{dbltopnumber}{1}
\makeatletter
\renewcommand{\cvprsect}[1]{\cvprsection[#1]{\texorpdfstring{\hskip -1em.~}{}#1}}
\renewcommand{\cvprsubsect}[1]{\cvprsubsection[#1]{\texorpdfstring{\hskip -1em.~}{}#1}}
\makeatother

\raggedbottom
\begingroup
\nolinenumbers
\hypersetup{linkcolor=black}
\maketitlesupplementary
\makeatletter
\def\numberline#1{\hb@xt@\@tempdima{#1.\hfil}}
\renewcommand*{\l@subsection}{\@dottedtocline{2}{1em}{2.5em}}
\patchcmd{\@dottedtocline}{\rightskip \@tocrmarg}
  {\rightskip \@tocrmarg plus 1fil}{}
  {\PackageError{supp}{Could not set contents spacing}{}}
\makeatother
\tableofcontents
\newpage
\endgroup

\appendix
\setcounter{figure}{0}
\setcounter{table}{0}
\setcounter{equation}{0}
\renewcommand{\thefigure}{S\arabic{figure}}
\renewcommand{\thetable}{S\arabic{table}}
\renewcommand{\theequation}{S\arabic{equation}}
\renewcommand{\theHfigure}{S\arabic{figure}}
\renewcommand{\theHtable}{S\arabic{table}}
\renewcommand{\theHequation}{S\arabic{equation}}
\section{Compute and comparison methods}
\label{sec:compute}

We describe the hardware, comparisons and records for \twsname{} (\tws{}); \cref{sec:sessions} reports individual sessions.

\subsection{Compute and editor configuration}

\paragraph{Relation to the weight-space programming report.}
\tws{} adapts Experiment 8 of the weight-space programming report~\cite{takehi2026sparks} to video grounding.
We rebuild its unpublished harness, the program that runs the conversation and the forward-pass tool, from the released first prompt and logs.
We use two settings and add black-frame measurements to both.
The attention-only setting allows 10 forward-pass calls and edits of attention weights only.
The report allows 10 to 50 calls (50 for its main result) and edits of all tensors except embeddings, output layer and final norm; the corresponding edit changes two MLPs.
The full-weight setting follows the report's budget of 50 calls (30 for runs 18 and 19), its editable weights, batch sizes and access to attention measurements.
Both settings use 480 practice questions, where the report uses about 15,000 practice problems. \Cref{tab:sessions} gives the editor and setting of every session.

\paragraph{Forward-pass calls and restrictions.}
In the attention-only setting, a \emph{generation call} answers up to 480 practice questions with the base or edited model, or 240 with both. Comparisons return paired mIoU differences and standard errors; activations are optional.
A \emph{teacher-forced call} scores up to 512 question--answer pairs and returns per-token log-probabilities, the top-$k$ next tokens ($k \le 50$) and residual-stream states.
The full-weight setting uses the report's batch sizes: up to 256 questions for a call with both models and 1,024 pairs for a teacher-forced call.
Either call can replace every frame with black to measure behaviour without visual evidence.
The attention-only setting permits changes to 144 tensors: the $q$, $k$, $v$ and $o$ projections in all 36 layers.
The tool that saves edits rejects changes that only rescale a whole tensor.
The CPU Python environment has read-only model files and no internet or GPU access, and it cannot run the model on inputs. All model measurements use the forward-pass tool.
The editor must diagnose a failure, find where in the model it arises by intervening at several strengths, and compare models on the same questions.

\paragraph{Practice check and test details.}
Predefined checks show that, for events in the final third of a practice video, Qwen3-VL-8B answers on average 39\% of the video length too early; the editor is not told this result.
Each baseline follows an evaluation rule fixed before it was run.
The test sets contain the visual VUE-TR and VUE-TR-V2~\cite{vidi2025,vidi2} questions whose videos we could download: 512 of 562 and 703 of 727, respectively.
The parser in the released TimeLens code~\cite{zhang2026timelens} reads two m:ss or h:mm:ss timestamps in those formats regardless of the unit word that follows.
Otherwise, it reads the first ``a - b'' range, then the first ``a to b'' range if needed, and finally the first two numbers, all in seconds.

\paragraph{Conversation and CPU execution.}
Claude Code uses maximum reasoning effort, disables built-in tools, and receives the system instruction ``Answer the user's request.''
Each turn explains the previous result, hypothesis, next step and possible refutation in English, followed by exactly one Python run or forward-pass call.
Every 40 turns, a fresh conversation receives the task, the notes the editor has written, its current edit and the remaining budgets; the main paper's ``one conversation with the editor'' means this whole session, which for Claude Opus 5.5 is refreshed in this way with only its notes, current edit and budgets.
A persistent CPU Python environment provides NumPy, SciPy, scikit-learn, PyTorch (CPU), pandas, SymPy, NetworkX, safetensors and tokenizers.
The editor may make at most 300 Python runs, each limited to 900\,s and 96\,GB of memory.
Our sandbox uses standard Linux process isolation in place of the original report's sandboxing tool.

\paragraph{Hardware and elapsed time.}
In runs 3 and 4, each editor session uses one CPU node and each forward-pass call uses one NVIDIA L40S GPU. In run 3, a single forward-pass call takes 10 to 47 minutes; \cref{sec:sessions} gives the total times of runs 3 and 4.
Test evaluations of runs 3 and 4, their base model, the Qwen3-VL-8B LoRA~\cite{hu2022lora} model and visual contrastive decoding use NVIDIA A30 GPUs.
The Qwen3.5-0.8B~\cite{qwen2026qwen35} test evaluations use the forward-pass tool on three L40S GPUs.
LoRA training takes 2 hours and 51 minutes on one L40S GPU.
Runs 10 to 15 use the main paper's second machine, which has four NVIDIA A40 GPUs and is called the A40 machine below. Each forward-pass call or test evaluation there uses one GPU.
Runs 16, 17, 27 and 28 use B200 GPUs with PyTorch's scaled dot-product attention for forward-pass calls and test evaluations, and the generation controls of \cref{sec:answer-controls} use V100 GPUs.
The steering edits are built and tested on the A40 machine, and every random control is tested on the same GPUs as the edit it is compared with.
Scores of the same weights differ slightly across GPU types, for example 29.54, 29.55 and 29.59 mIoU for the TimeLens-8B base model on A40, B200 and V100 GPUs, so every comparison pairs answers from one GPU type.

\paragraph{Audit of the editor configuration.}
GPT-6 Astra's command line, disabled tools, one-line instruction, model and reasoning effort match the report's released records. No built-in tool runs in any turn we checked.
Median reasoning lengths range from 63 to 516 tokens per turn across our sessions and from 192 to 252 across the report's three runs.
The report's medians use changes in its reasoning-token counter, excluding turns without an increase.

\Needspace{0.75\textheight}
\subsection{Separator diagnostics}
\label{sec:separator-diagnostics}

\begin{figure}[!htb]
\centering
\includegraphics[width=\linewidth]{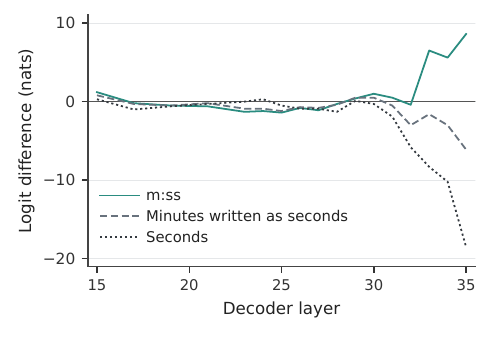}
\caption{\textbf{The separator preference changes in the final decoder layers.} Mean logit difference between ``:'' and ``.'', read through the final norm and output layer of Qwen3-VL-8B. Development questions on videos of at least 30 minutes are grouped by the base answer's format. Positive values favour a colon; negative values favour a decimal point. The curves separate in the final layers. Main-paper Secs.~4.2 and~4.3 describe both repairs.}
\label{fig:editor-diagnostics}
\end{figure}

The logit lens reads each decoder layer's residual stream through the final norm and output layer. We measure the mean difference between the logits for ``:'' and ``.'' after each of the 36 layers; \cref{fig:editor-diagnostics} shows layers 15 to 35. Development questions on videos of at least 30 minutes are grouped by the base answer's format: m:ss, minutes written as seconds, or seconds. All three groups stay near zero until layer 31. From layer 32 to layer 33, the mean logit difference increases by about 7 nats for m:ss answers but only 1.4 for minutes written as seconds; the final layer widens the gap between the groups.
On 32 answers with minutes written as seconds, a teacher-forced call with the query edit at $\kappa=2$ raises the log-odds of ``:'' over ``.'' from $-5.67$ to $-2.82$.
For the held-out example in main-paper Fig.~1, both answers begin ``\texttt{The event happens in 12}''. The base model continues with a digit and writes ``\texttt{124.0 - 158.0 seconds}'' (2:04 to 2:38); the query edit writes a colon and gives 12:27 to 14:12. Main-paper Secs.~4.2 and~4.3 give the query and value/output constructions and their submitted strengths.

\subsection{Comparison methods}

\paragraph{Global rescaling and a trained grounding model.}
For Qwen3-VL-8B, fitting one rescaling factor $c$ on practice predictions selects $c = 1.0$, leaving only clipping to video bounds. The test difference is not significant: $+0.09$ $[0.00, +0.20]$.
TimeLens2-8B, trained for grounding, uses its own prompt on the same questions and machines.
It scores 49.92 on VUE-TR and 40.92 on VUE-TR-V2, $+19.69$ above the Qwen3-VL-8B base model on the pooled test sets.

\paragraph{Visual contrastive decoding with black frames.}
Our visual contrastive decoding variant~\cite{leng2024vcd} uses black frames instead of noise-distorted images and reduces the contrast weight as confidence increases, where the original keeps it fixed. Greedy decoding combines the base model's real-video and black-video logit streams into one answer.
At step $t$, let $p_t$ be the highest next-token probability with the real-video input.
We consider tokens with probability at least a tenth of $p_t$ and choose the one with the largest real-video logit minus $\alpha_t$ times its black-video logit, using the same prompt in both streams.
The weight $\alpha_t = 1/(1 + e^{(p_t - 0.5)/0.1})$ falls from about 1 to about 0 as $p_t$ increases.
With Qwen3-VL-8B-Instruct~\cite{bai2025qwen3vl}, this method scores 31.44 on VUE-TR and 23.89 on VUE-TR-V2, a gain of $+2.05$ $[+1.13, +3.00]$ over the base model.
Neither the query nor the value/output edit differs significantly from it ($-0.75$ $[-1.74, +0.28]$ and $+0.83$ $[-0.33, +2.03]$, respectively). Unlike weight editing, it requires a second decoding stream.

\paragraph{Difference-of-means steering through weight edits.}
For Qwen3-VL-8B, Qwen3-VL-2B and TimeLens-8B, we construct the steering edit of main-paper Sec.~5 from the diagnosed timestamp decision, without an LLM.
At layer $\ell$, $u_\ell$ is the unit direction of the difference between the mean residual-stream states of the two answer types at that decision position.
We apply $W' = W + \alpha\, u_\ell u_\ell^{\top} W$ to the attention and MLP output projections in five consecutive layers.
We test two five-layer windows per model: the window that best separates the answer types and the window that contains the sessions' edits. Each uses $\alpha \in \{-1, 1, 2, 4\}$.
Under predefined rules, the best development candidate is checked once on held-out questions, then tested once regardless of that check.

For Qwen3-VL-8B, the decision is the separator after the first number, ``\texttt{:}'' or ``\texttt{.}''.
The two types occur in 233 development answers (133 and 100), and the two layer windows coincide.
Qwen3-VL-2B also writes minutes as seconds, so we use the same separator decision, with 234 development answers (33 and 201).
For TimeLens-8B, the decision is whether a fourth integer digit follows the third.
Its own answers contain this continuation only 4 times, so we use states from reference answers written in the model's format: 95 positions followed by another integer digit and 367 followed by a decimal point.
\Cref{tab:steering} reports every candidate.

\begin{table*}[t]
\centering
\caption{\textbf{Difference-of-means steering built from each diagnosis.} (a) Development $\Delta$mIoU over the base model for every five-layer window and strength $\alpha$. Bold marks the selected candidate, evaluated on held-out practice questions (mean $\pm$ one standard error) and the test sets. $\dagger$ marks the window that best separates the answer types; $\ddagger$ marks the window containing the sessions' edited layers. These coincide for Qwen3-VL-8B. (b) Each session's edit minus the selected steering candidate; bold marks positive gains whose intervals exclude zero. All test gains pool 1,215 questions and use 95\% bootstrap confidence intervals over videos. Development and held-out counts reuse the split of run 3 (234 and 246) for both Qwen3-VL models and the split of run 15 (239 and 241) for TimeLens-8B. All answers come from the A40 machine.}
\label{tab:steering}
\small
\setlength{\tabcolsep}{3pt}
\begin{tabular*}{\textwidth}{@{\extracolsep{\fill}}llrrrrrrr@{}}
\toprule
\multicolumn{9}{@{}l}{\textbf{(a) Steering candidates and evaluation of the selected candidate}} \\
\addlinespace[3pt]
 & & \multicolumn{4}{c}{Development $\Delta$mIoU} & \multicolumn{3}{c}{Selected candidate} \\
\cmidrule(lr){3-6}\cmidrule(l){7-9}
Base model & Layers & $\alpha=-1$ & $\alpha=1$ & $\alpha=2$ & $\alpha=4$ & Held-out & Test & 95\% CI \\
\midrule
Qwen3-VL-8B & 31--35$^{\dagger\ddagger}$ & $-9.51$ & $+4.33$ & $\mathbf{+5.14}$ & $+0.25$ & $+5.04 \pm 1.51$ & $+2.07$ & $[+0.92, +3.27]$ \\
\addlinespace[2pt]
Qwen3-VL-2B & 22--26$^{\dagger}$ & $-9.30$ & $\mathbf{+2.64}$ & $-0.50$ & $-15.45$ & $-0.12 \pm 0.90$ & $+0.87$ & $[+0.08, +1.74]$ \\
 & 16--20$^{\ddagger}$ & $-7.85$ & $-6.48$ & $-16.23$ & $-19.24$ & & & \\
\addlinespace[2pt]
TimeLens-8B & 18--22$^{\dagger}$ & $-6.41$ & $+4.52$ & $\mathbf{+7.84}$ & $+3.54$ & $+5.46 \pm 1.22$ & $+4.79$ & $[+3.45, +6.23]$ \\
 & 30--34$^{\ddagger}$ & $-4.19$ & $+0.46$ & $-1.96$ & $-15.44$ & & & \\
\bottomrule
\end{tabular*}

\vspace{6pt}
\begin{tabular*}{\textwidth}{@{\extracolsep{\fill}}rrlrrlrrl@{}}
\toprule
\multicolumn{9}{@{}l}{\textbf{(b) Test $\Delta$mIoU: session edit minus steering}} \\
\addlinespace[3pt]
\multicolumn{3}{c}{Qwen3-VL-8B} & \multicolumn{3}{c}{Qwen3-VL-2B} & \multicolumn{3}{c}{TimeLens-8B} \\
\cmidrule(r){1-3}\cmidrule(lr){4-6}\cmidrule(l){7-9}
Run & $\Delta$mIoU & 95\% CI & Run & $\Delta$mIoU & 95\% CI & Run & $\Delta$mIoU & 95\% CI \\
\midrule
3 & $-0.16$ & $[-1.26, +0.92]$ & 12 & $\mathbf{+1.24}$ & $[+0.21, +2.27]$ & 15 & $+0.82$ & $[-0.55, +2.14]$ \\
4 & $+0.78$ & $[-0.25, +1.84]$ & 14 & $+0.61$ & $[-0.40, +1.59]$ & 16 & $-1.54$ & $[-2.85, -0.31]$ \\
 & & & & & & 17 & $\mathbf{+3.03}$ & $[+1.78, +4.31]$ \\
\bottomrule
\end{tabular*}
\end{table*}

\paragraph{Asking for the time format.}
Three prompts test whether specifying the time format repairs the errors. Each retains the standard prompt's first sentence and replaces its format line.
The \emph{clock} prompt requests hours, minutes and seconds: ``\texttt{The event happens in <start HH:MM:SS> - <end HH:MM:SS>}''.
The \emph{seconds} prompt requests total seconds from the video's start and gives a worked example: ``a moment 12 minutes 34 seconds into the video is 754 seconds''.
The \emph{minutes} prompt requests both times as m:ss, the format favoured by the Qwen3-VL-8B edits, with examples 12:34 and 75:05.
The answer parser handles seconds, m:ss and h:mm:ss.

We compare the clock and seconds prompts on the 120 screening questions of \cref{sec:sessions}.
Under predefined rules, the better prompt is tested once with the base weights (\cref{tab:prompt}).
The minutes prompt has its own rule, fixed before it was tested: it is tested only if its screening score exceeds the selected prompt's score, which happens only for Qwen3-VL-8B.
For that model, the clock prompt moves the unit error rather than removing it: an event at 29:04 receives ``\texttt{29:13:00}'', read as 29 hours.
The seconds prompt lowers test mIoU.
The minutes prompt produces m:ss answers and improves on the screening questions, but lowers test mIoU overall. Its difference on videos of at least 30 minutes is not significant (\cref{tab:prompt}).
TimeLens-8B ignores the minutes request and continues to write seconds.
The seconds prompt improves overall but not significantly on videos of at least 30 minutes, where the dropped digit occurs.
Every session edit and steering edit scores above the tested prompts.

\begin{table*}[t]
\centering
\caption{\textbf{Prompts that specify the time format.} Left: each base model's mIoU with four prompts on the 120 practice screening questions. Right: each tested prompt's gain over the standard-prompt base model on 1,215 test questions, paired per question, followed by each edit's gain over that prompt. All results use the A40 machine, so the standard-prompt scores differ slightly from the screening scores in \cref{sec:sessions} (21.25 and 29.24); intervals are 95\% bootstrap intervals over videos. Session edits are listed as query then value/output for Qwen3-VL-8B, and runs 15, 16 and 17 for TimeLens-8B.}
\label{tab:prompt}
\footnotesize
\setlength{\tabcolsep}{3.5pt}
\sbox0{\begin{tabular}{lrrrrlrrrr}
\toprule
 & \multicolumn{4}{c}{Screening mIoU with each prompt} & & \multicolumn{2}{c}{Prompt minus standard} & \multicolumn{2}{c}{Edit minus prompt} \\
\cmidrule(lr){2-5}\cmidrule(lr){7-8}\cmidrule(lr){9-10}
Base model & Standard & Clock & Seconds & Minutes & Tested prompt & All 1,215 & $\geq$30 min & Sessions' edits & Steering \\
\midrule
Qwen3-VL-8B & 20.92 & 3.51 & 12.52 & 22.66 & Seconds & $-1.86$ {\scriptsize$[-3.28, -0.52]$} & $-2.78$ {\scriptsize$[-4.57, -0.96]$} & $+3.77$, $+4.71$ & $+3.93$ \\
 & & & & & Minutes & $-3.77$ {\scriptsize$[-5.48, -2.01]$} & $+0.94$ {\scriptsize$[-1.54, +3.24]$} & $+5.68$, $+6.62$ & $+5.84$ \\
\addlinespace
TimeLens-8B & 29.51 & 28.38 & 29.73 & 27.99 & Seconds & $+1.04$ {\scriptsize$[+0.03, +2.06]$} & $+0.92$ {\scriptsize$[-0.65, +2.48]$} & $+4.57$, $+2.22$, $+6.78$ & $+3.76$ \\
\bottomrule
\end{tabular}}%
\ifdim\wd0>\textwidth \resizebox{\textwidth}{!}{\usebox0}\else\usebox0\fi
\end{table*}

\paragraph{Adding the edits to LoRA.}
We add each unchanged Qwen3-VL-8B edit to the LoRA model: affected tensors become the edited tensor plus its LoRA change. Each combination is tested once against the LoRA model alone, with the merged LoRA update in both models.
On the A40 machine, LoRA gains $+13.42$ $[+11.42, +15.44]$ over the base model.
Neither edit provides a significant further gain: query edit plus LoRA minus LoRA gives $-0.06$ $[-0.44, +0.33]$, and value/output edit plus LoRA minus LoRA gives $-0.04$ $[-0.36, +0.28]$.
Differences on videos of at least 30 minutes are also not significant ($+0.44$ and $+0.18$).

\paragraph{An editor that sees only the answers.}
One Claude Opus 5.5 session per model receives the 480 practice questions as answered on the A40 machine: queries, video lengths, base answers and true intervals, with anonymous video labels.
It also receives the test answer parser, IoU scoring and Python, but no model, video or other file.
Each session takes about 15 minutes to diagnose errors and write a correction rule. Under predefined rules, this is applied once, unchanged, to all 1,215 test answers.

For Qwen3-VL-8B, the editor finds minutes written as seconds in 40 practice answers and tenths of a minute written as seconds in 25, such as 118.0 for 11.8 minutes.
Let $D$ be the video duration and $[s,e]$ the first parsed interval, all in seconds.
The rule swaps reversed endpoints and sets $e=s+1$ when they are equal; if no pair is readable, it returns the whole video.
A plain-seconds answer with integer endpoints, $5\le s<15$, $e-s\le8$ and $60s+20<D$ becomes $[60s+20,60s+120]$.
Otherwise, a plain-seconds answer with $80\le s<130$ and $6s+10<D$ becomes $[6s+10,6e+10]$.
Clock-format answers (m:ss or h:mm:ss) become $[s-1,e+4]$.
After correction, an interval starting at or beyond $D$ is moved to end at $D$, retaining its length up to $D$. The rule then clips endpoints to the video and discards empty intervals.

For TimeLens-8B, the editor finds times ten times too small in 135 of 265 answers on videos of about 1,300 seconds or more.
Its rule uses the evaluation parser's first interval. If parsing fails, it swaps reversed endpoints in the first time pair or adds one second when they are equal; no readable pair means no interval.
It maps $[s,e]$ to $[10s+5,\min(D,10e+5)]$ only when $D\ge1{,}300$, $s>0$, $10e\le D$ and $10(e-s)<0.12D$.
If clipping makes this interval empty, it omits the five-second offset.
When the transformation does not apply, the interval is unchanged except that its end is clipped to $D$ if $s<D<e$.
Neither rule uses the query. Practice mIoU rises from 21.55 to 23.96 for Qwen3-VL-8B and from 30.32 to 41.50 for TimeLens-8B.

On the test sets, the Qwen3-VL-8B rule's gain is not significant overall ($+0.56$ $[-0.05, +1.18]$) but is significant on videos of at least 30 minutes ($+1.21$ $[+0.37, +2.10]$).
The query edit, value/output edit and steering exceed it by $+1.35$ $[+0.39, +2.35]$, $+2.29$ $[+1.21, +3.41]$ and $+1.51$ $[+0.30, +2.81]$, respectively.
The TimeLens-8B rule gains $+6.04$ $[+4.65, +7.48]$, compared with $+5.55$ for the times-ten rule, and gains $+17.65$ on videos of at least 30 minutes.
Run 17 exceeds it by $+1.79$ $[+0.41, +3.16]$.
Run 15 and steering do not differ significantly from it; run 16 scores lower ($-2.78$ $[-4.15, -1.51]$).
Answers alone reveal both errors, but model access also locates them, reveals encoded times absent from the answer (\cref{sec:supp-separator}), and enables repairs that outperform the best rule written from answers.

\paragraph{Side effects on decimals and times.}
Because the edits affect punctuation and decimal points, we test ordinary decimals and times using text alone.
Two hundred generated questions (seed 0), 50 per task, request sums of two decimals, conversions from seconds to m:ss, conversions from m:ss to seconds, and division of a whole number by ten.
We use greedy decoding and mark an answer correct if its first number or m:ss value matches the reference answer.
Qwen3-VL-8B answers 92.5\% correctly.
Each edit leaves 199 of the 200 answers unchanged word for word, with accuracies of 92.0\% and 92.5\%.
TimeLens-8B answers 80.0\% correctly.
Runs 15 and 17 do not differ significantly from it (81.5\% and 82.0\%), but run 16 falls to 56.5\%: a change of $-23.5$ $[-30.0, -17.5]$ points, with bootstrapping over questions.
Run 16 drops decimal points, writing ``\texttt{67590}'' for 675.9, and answers only 4\% of divisions by ten correctly, compared with 92\% for the base model.
Run 17's selector, the direction that decides when its edit acts, is made orthogonal to states at 24 correct decimal points in practice answers (\cref{sec:tl8-sessions}); run 17 shows no such loss.
These text-only tests do not assess video question answering; \cref{sec:locality} reports video question answering and other video tasks.

\subsection{Materials to be released}

The camera-ready release will include the harness, all session records (\cref{sec:sessions}), the submitted edits and the scoring code.
The harness contains the program that runs the conversation, the sandboxed Python environment, the forward-pass tool and the code that sends forward-pass calls to the GPUs.
The records contain every prompt and reply, editor explanation, Python run, forward-pass call and final answer, together with the practice splits and editor model versions.
The release will also list all 1,289 visual VUE-TR and VUE-TR-V2 test questions, marking the 1,215 we evaluate and the 74 whose videos could not be downloaded.

\section{Editor sessions by model and setting}
\label{sec:sessions}

\Cref{tab:sessions} reports practice and pooled test results for the 21 sessions of the main paper, runs 1 to 19, 21 and 22, including unsuccessful and stopped runs. Run numbers match the session records, which will be released with the harness (\cref{sec:compute}); runs 3 and 4 are the main paper's query and value/output edits. Runs 27 and 28 repeat the TimeLens-8B setting (\cref{sec:new-sessions}). A session's final edit is tested even when the editor calls its own held-out check inconclusive; a session submits nothing only when the editor restores the base weights or the session stops. The descriptions below explain each edit and give results beyond the table.

Claude Opus 5.5 runs through Claude Code and GPT-6 Astra through Codex, both at maximum reasoning effort with built-in tools disabled.
For the primary result, each submitted edit is evaluated once against the base model on all 1,215 VUE-TR and VUE-TR-V2 questions (main-paper Sec.~3). Additional comparisons below use the same fixed edits. A 95\% bootstrap interval over videos above or below zero indicates improvement or degradation; an interval containing zero is not significant, not evidence of equivalence.
\Cref{sec:compute} gives the hardware; each comparison pairs answers from the same GPUs.

\begin{table*}[t]
\centering
\caption{\textbf{The 21 editing sessions of the main paper, including unsuccessful and stopped runs.} Run numbers match the session records. Claude and Astra are Claude Opus 5.5 and GPT-6 Astra. Attn. and Full allow attention-only and full-weight edits, with budgets of 10 and 50 forward-pass calls (30 for runs 18 and 19); Calls counts the calls used. Full excludes embeddings, output layer and final norm; batch sizes and access to attention measurements follow the weight-space programming report. Run 1 edits only LoRA tensors and is compared with TimeLens2-8B without the adapter. Query updates are rank one per head; $o$ denotes the attention output projection and ``think.'' thinking mode. Practice $\Delta$mIoU gives paired differences with one standard error on $n$ questions. Test $\Delta$mIoU pools the 1,215 VUE-TR and VUE-TR-V2 questions, with 95\% bootstrap intervals over videos. Bold marks intervals above zero; run 2's is below zero. A dash in the test column means no edit was submitted. $^{a}$ 478 practice questions, none held out. $^{b}$ Base answers were inspected before selecting run 2's direction. $^{c}$ Run 4 reused its held-out questions to compare three strengths.}
\label{tab:sessions}
\footnotesize
\setlength{\tabcolsep}{3.2pt}
\resizebox{\textwidth}{!}{%
\begin{tabular}{rlllllrrrrrl}
\toprule
 & & & & \multicolumn{3}{c}{Submitted edit} & & & \multicolumn{2}{c}{Practice} & \\
\cmidrule(lr){5-7}\cmidrule(lr){10-11}
Run & Editor & Base model & Access & Weights & Layers & Tensors & Turns & Calls & $\Delta$mIoU & $n$ & Test $\Delta$mIoU \\
\midrule
1 & Claude & TimeLens2-8B + LoRA & LoRA & top LoRA component removed & adapter & 120 & 139 & 10 & $+0.13 \pm 0.57^{a}$ & 478 & $-0.24$ {\scriptsize$[-0.87, +0.37]$} \\
2 & Claude & TimeLens2-8B & Attn. & attention output, rank 1 & 30, 32--34 & 4 & 258 & 10 & $+0.23 \pm 0.45^{b}$ & 240 & $-1.02$ {\scriptsize$[-1.54, -0.55]$} \\
3 & Claude & Qwen3-VL-8B & Attn. & query & 31--35 & 5 & 189 & 10 & $+3.77 \pm 1.11$ & 246 & $\mathbf{+1.30}$ {\scriptsize$[+0.63, +2.01]$} \\
4 & Claude & Qwen3-VL-8B & Attn. & value and output & 35 & 2 & 277 & 10 & $+3.96 \pm 1.37^{c}$ & 246 & $\mathbf{+2.88}$ {\scriptsize$[+1.85, +3.97]$} \\
5 & Astra & Qwen3-VL-8B & Attn. & none & -- & -- & 28 & 10 & $-0.61 \pm 0.79$ & 240 & -- \\
6 & Claude & Qwen3-VL-8B & Full & none (stopped) & -- & -- & 24 & 1 & -- & -- & -- \\
7 & Astra & Qwen3-VL-8B & Full & 7 heads of $o$, scaled by 1.5 & 21, 23, 24 & 3 & 117 & 50 & $+0.57 \pm 1.24$ & 239 & $-0.20$ {\scriptsize$[-1.13, +0.67]$} \\
8 & Astra & Qwen3.5-0.8B & Full & 1 head of $o$ & 15 & 1 & 112 & 50 & $+2.47 \pm 1.24$ & 254 & $+0.36$ {\scriptsize$[-0.62, +1.40]$} \\
9 & Astra & Qwen3-VL-2B & Full & 128 MLP output columns & 25 & 1 & 108 & 50 & $+1.31 \pm 1.27$ & 240 & $-0.12$ {\scriptsize$[-1.04, +0.82]$} \\
10 & Astra & Qwen3-VL-8B & Attn. & attention output, rank 1 & 24 & 1 & 25 & 10 & $+2.39 \pm 1.10$ & 240 & $+0.28$ {\scriptsize$[-0.62, +1.15]$} \\
11 & Astra & Qwen3-VL-8B & Full & none (stopped) & -- & -- & 35 & 13 & -- & -- & -- \\
12 & Astra & Qwen3-VL-2B & Full & 2 heads of $o$ & 20, 16 & 2 & 110 & 50 & $+4.40 \pm 1.27$ & 170 & $\mathbf{+2.11}$ {\scriptsize$[+1.36, +2.89]$} \\
13 & Astra & Qwen3-VL-8B & Attn. & 3 query heads & 21, 22 & 2 & 30 & 10 & $+0.09 \pm 0.56$ & 240 & $+0.36$ {\scriptsize$[-0.14, +0.87]$} \\
14 & Astra & Qwen3-VL-2B & Full & 2 heads of $o$ & 19 & 1 & 117 & 50 & $+1.02 \pm 1.08$ & 240 & $\mathbf{+1.49}$ {\scriptsize$[+0.63, +2.40]$} \\
15 & Astra & TimeLens-8B & Full & MLP output, rank 1 & 34 & 1 & 112 & 50 & $+7.32 \pm 1.27$ & 241 & $\mathbf{+5.61}$ {\scriptsize$[+4.35, +7.02]$} \\
16 & Astra & TimeLens-8B & Full & MLP output, rank 1 & 34 & 1 & 127 & 50 & $+6.30 \pm 1.71$ & 137 & $\mathbf{+3.13}$ {\scriptsize$[+2.19, +4.14]$} \\
17 & Astra & TimeLens-8B & Full & MLP output, rank 1 & 34 & 1 & 120 & 50 & $+10.72 \pm 1.60$ & 240 & $\mathbf{+7.89}$ {\scriptsize$[+6.28, +9.46]$} \\
18 & Astra & Qwen3.5-0.8B (think.) & Full & direction removed from $o$ & 16--23 & 8 & 68 & 30 & $+3.95 \pm 1.79$ & 276 & $\mathbf{+2.33}$ {\scriptsize$[+0.88, +3.79]$} \\
19 & Astra & Qwen3.5-0.8B (think.) & Full & 1 head of $o$, halved & 15 & 1 & 64 & 30 & $-0.85 \pm 1.59$ & 180 & $+1.24$ {\scriptsize$[-0.14, +2.70]$} \\
21 & Astra & InternVL3.5-8B & Full & \begin{tabular}[t]{@{}l@{}}32 directions removed\\from 1 head of $o$\end{tabular} & 23 & 1 & 128 & 50 & $+1.00 \pm 0.45$ & 230 & $+0.11$ {\scriptsize$[-0.33, +0.49]$} \\
22 & Astra & MiMo-VL-7B-SFT-2508 & Full & none (stopped) & -- & -- & 78 & 32 & -- & -- & -- \\
\bottomrule
\end{tabular}}
\end{table*}

\subsection{TimeLens2-8B sessions}

Runs 1 and 2 use the grounding model TimeLens2-8B~\cite{zhu2026timelens2}.
Two predefined checks on 480 practice questions about videos of at least 10 minutes find no systematic failure: late events are localised as well as early ones, and one correction based on video length does not help.
These practice questions come from TimeLens2-93K, the training data of TimeLens2-8B, so the checks may understate its errors.
When a video is given queries from other videos, the answer stays the same for only 7.9\% of videos, so the answers follow the query.
The other sessions therefore use Qwen models, which place late events too early, and TimeLens-8B, whose screening shows a large correctable error (\cref{sec:screening}).

\paragraph{Run 1 repairs an adapter-induced format error.}
Run 1 starts from a LoRA adapter for TimeLens2-8B that was trained for a different task; merged into TimeLens2-8B, it breaks the generated answer format.
Claude Opus 5.5 traces this error to the top singular components of the adapter's value and output updates. It removes the top component from 120 of the adapter's 144 LoRA $B$ matrices: those of the query, key, value and output projections in layers 0 to 23 and of the query and key projections in layers 24 to 35.
On all practice questions, none held out, this gains $+20.13 \pm 1.71$ over the trained adapter. The repair restores the format but does not significantly improve on TimeLens2-8B without the adapter, on practice or test questions (\cref{tab:sessions}).

\paragraph{Run 2 suppresses useful extra spans.}
Claude Opus 5.5 has attention-only access to TimeLens2-8B's 144 attention tensors, without an adapter.
The main error is choosing the wrong interval, and Claude Opus 5.5 finds no component that repairs it. Instead, it multiplies by seven an output component of six heads in layers 30 and 32 to 34 that encourages the answer to end after its first span.
The small held-out practice gain becomes a test loss (\cref{tab:sessions}), entirely from the 313 questions with at least two true spans ($-3.85$ $[-5.51, -2.31]$). The base model's additional spans are mostly correct.

\subsection{Qwen3-VL-8B in the attention-only setting}

\begin{figure}[t]
\centering
\includegraphics[width=\linewidth]{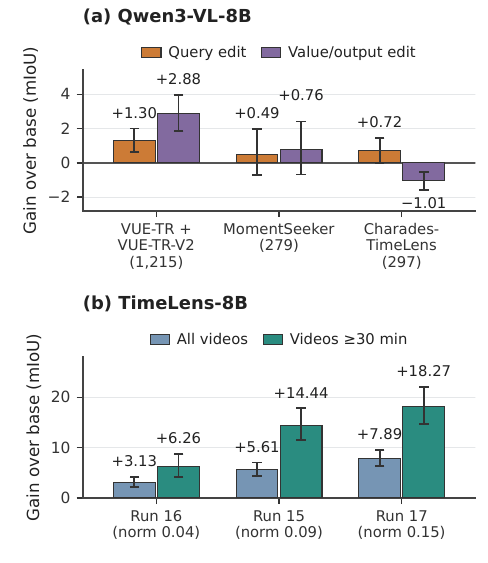}
\caption{\textbf{Where the weight edits generalise.} Bars show paired test gains over the base model, with 95\% bootstrap confidence intervals over videos. (a) Both Qwen3-VL-8B edits improve on the pooled VUE-TR and VUE-TR-V2 test sets (1,215 questions). Neither differs significantly from the base model on MomentSeeker (279), while the value/output edit reduces accuracy on short Charades-TimeLens videos (297). (b) The three TimeLens-8B edits gain more on the 422 questions from videos of at least 30 minutes than on all 1,215 questions. Norm is the weight change's Frobenius norm divided by the base tensor's norm. Runs are independent editing sessions ordered by this norm, not strengths of one edit. The subsets overlap, so their gains are not independent comparisons.}
\label{fig:transfer}
\end{figure}

\paragraph{Run 3 makes the query edit.}
The session takes 11 hours and 31 minutes across five conversations and uses 172 of 300 Python runs.
Claude Opus 5.5 identifies minutes written as seconds and applies the rank-1 query update of main-paper Sec.~4.2 to layers 31 to 35; \cref{tab:base_practice} gives its strengths.
Differences from the base model are not significant on MomentSeeker~\cite{yuan2025momentseeker} ($7.26 \to 7.76$, $+0.49$ $[-0.71, +1.96]$) or Charades-TimeLens~\cite{gao2017tall,zhang2026timelens} ($38.97 \to 39.70$, $+0.72$ $[-0.04, +1.45]$; \cref{fig:transfer}).

After the session, one random draw (seed 0) gives rank-1 controls matching the edit's size in each head of the same five tensors.
On all 480 practice questions, a random query shift and probe lower mIoU by $3.05 \pm 0.66$; the edit's query shifts with a random probe lower it by $1.61 \pm 0.67$; and random query shifts with the edit's probe lower it by $6.43 \pm 0.93$.
On the A40 machine, their test differences are not significant or negative ($-0.04$, $-0.02$, $-2.66$); the query edit exceeds them by $+1.95$, $+1.93$ and $+4.58$, respectively.
Each control is compared only with its own edit.

\begin{table}[t]
\centering
\caption{\textbf{Edit-strength ablation for the query edit (main-paper Sec.~4.2).} $\Delta$ denotes edited minus base mIoU, with one standard error. Development (dev.) questions fit the edit and select $\kappa$. Held-out questions are excluded from both choices. The negative strength is a sign-reversal control.}
\label{tab:base_practice}
\small
\setlength{\tabcolsep}{2pt}
\begin{tabular}{llc}
\toprule
Edit strength & Practice subset & $\Delta$mIoU \\
\midrule
$\kappa = 4$ & 234 dev. & $+1.90 \pm 0.98$ \\
\textbf{$\boldsymbol{\kappa} = 4$ (submitted)} & \textbf{246 held-out} & $\mathbf{+3.77 \pm 1.11}$ \\
$\kappa = 4$ & all 480 & $+2.86 \pm 0.74$ \\
$\kappa = 3$ & all 480 & $+2.34 \pm 0.67$ \\
$\kappa = 6$ & 234 dev. & $+1.56 \pm 1.32$ \\
$\kappa = -4$ (sign reversed) & all 480 & $-7.04 \pm 0.98$ \\
\bottomrule
\end{tabular}
\end{table}

\paragraph{Run 4 makes the value/output edit.}
This independent repetition uses the same task, practice questions and budgets. It takes 13 hours and 57 minutes across seven conversations and uses 257 of 300 Python runs.
Claude Opus 5.5 reaches the same diagnosis but edits only layer 35's value and output projections. The added output vector raises the ``\texttt{:}'' logit independently of the attention pattern.
The practice questions reported in \cref{tab:sessions} were also used to compare three strengths.
The MomentSeeker difference is not significant ($+0.76$ $[-0.68, +2.41]$), while Charades-TimeLens mIoU changes by $-1.01$ $[-1.58, -0.51]$ (main-paper Sec.~5).
On the A40 machine, a value/output control preserves the rank and singular values but randomises the singular vectors. Its gain is not significant ($+0.08$); the edit exceeds it by $+2.77$.
\Cref{tab:base_test} gives both edits' test scores by dataset.

\begin{figure*}[t]
\centering
\includegraphics[width=\textwidth]{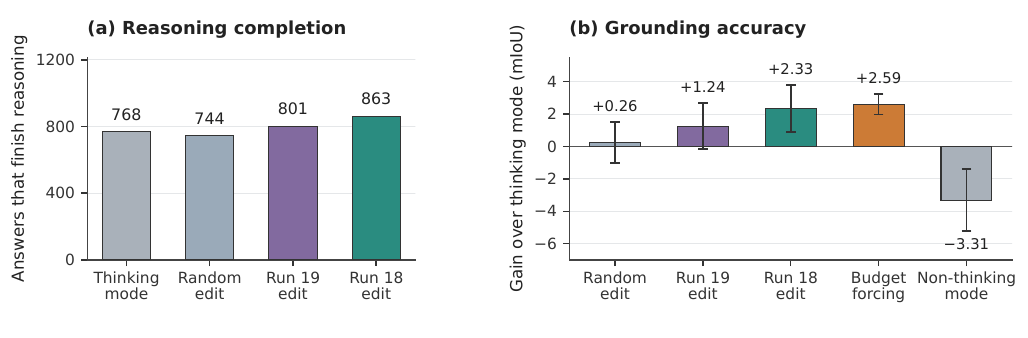}
\caption{\textbf{Run 18 finishes more answers but does not beat budget forcing.} All bars use the same 1,215 Qwen3.5-0.8B test questions. (a) Answers that finish reasoning within 1,024 tokens. Removing the repetition direction in run 18 raises the count from 768 to 863. Its random-direction control finishes 744 answers; halving one head's output in run 19 finishes 801. Budget forcing closes unfinished reasoning through a decoding rule, and non-thinking mode disables reasoning, so neither has a count of answers that finish reasoning on their own. (b) Paired mIoU gains over thinking mode, with 95\% bootstrap confidence intervals over videos. Run 18 improves over thinking mode and its random-direction control (the latter not significant after Holm correction, \cref{sec:holm}), but does not differ significantly from budget forcing ($-0.27$ $[-1.82, +1.26]$). More completed answers therefore do not establish an accuracy advantage over that baseline.}
\label{fig:reasoning}
\end{figure*}

\begin{table}[t]
\centering
\caption{\textbf{Qwen3-VL-8B edits and baselines on A30 GPUs, by dataset.} Qwen3-VL-8B-Instruct uses 448 evenly spaced frames. Entries are mIoU. The query edit changes 5 tensors and the value/output edit 2. Pooled paired gains have 95\% bootstrap intervals over videos. The output rule is the answer-side correction of main-paper Sec.~5. The LoRA model is fine-tuned on the editors' 480 practice questions.}
\label{tab:base_test}
\footnotesize
\setlength{\tabcolsep}{3pt}
\sbox0{\begin{tabular}{lrrr}
\toprule
Method & VUE-TR & VUE-TR-V2 & Pooled $\Delta$mIoU \\
\midrule
Base model & 29.95 & 21.43 & \\
Query edit & 31.24 & 22.74 & $+1.30$ {\scriptsize$[+0.63, +2.01]$} \\
Value/output edit & 33.36 & 23.93 & $+2.88$ {\scriptsize$[+1.85, +3.97]$} \\
Output rule & 30.66 & 21.76 & $+0.49$ {\scriptsize$[-0.07, +1.04]$} \\
LoRA model & 42.55 & 34.67 & $+12.97$ {\scriptsize$[+11.01, +15.00]$} \\
\bottomrule
\end{tabular}}%
\ifdim\wd0>\columnwidth \resizebox{\columnwidth}{!}{\usebox0}\else\usebox0\fi
\end{table}

\paragraph{Run 5 submits no edit.}
Under the same protocol, GPT-6 Astra recognises minutes written as seconds but expects to repair few answers. It instead investigates how video evidence affects the generated times.
Its best candidate doubles an attention-output direction associated with event location in layers 20 to 23.
Its $+0.76 \pm 0.74$ gain on 240 development questions does not survive the held-out check (\cref{tab:sessions}). GPT-6 Astra restores the base weights and submits no edit, so no test evaluation is permitted.

\paragraph{Runs 10 and 13 repeat run 5.}
Both sessions start run 5's setting from scratch.
Run 10 compares real and black frames to investigate answers placed too early. It amplifies one attention-output direction in layer 24 fivefold, $W' = W + 4\,u u^{\top} W$.
Its gain on held-out questions from unseen videos does not transfer to the test sets.
Run 13 attributes the errors to location rather than format. It doubles the dependence of each of three query heads on a direction: head 26 in layer 21 and heads 2 and 11 in layer 22.
GPT-6 Astra judges its held-out gain inconclusive. Neither session improves significantly on the test sets or repairs the minutes written as seconds that Claude Opus 5.5 found.

\subsection{Qwen3-VL-8B in the full-weight setting}

\paragraph{Run 6 stops after one measurement.}
Claude Opus 5.5 can edit 744 of Qwen3-VL-8B's 750 tensors.
We stopped run 6 after 20 Python runs and one forward-pass call so that the full-weight setting would use the report's editor, GPT-6 Astra.
The single call evaluates the base model on all 480 practice questions (21.7 mIoU). No edit is submitted or tested.

\paragraph{Run 7 amplifies seven attention heads.}
GPT-6 Astra receives run 6's setting.
It multiplies the outputs of seven heads by 1.5: heads 8, 10 and 11 in layer 21; heads 4, 5 and 31 in layer 23; and head 29 in layer 24.
It gains $+2.00 \pm 0.95$ on development questions, but GPT-6 Astra judges its held-out result, from 110 videos, inconclusive. Halving these heads instead loses $5.79 \pm 1.35$.
The test difference is also not significant.
Both Claude Opus 5.5 edits outperform run 7: the query edit by $+1.50$ $[+0.51, +2.61]$ and the value/output edit by $+3.08$ $[+1.96, +4.26]$.

\paragraph{Run 11 stops before submitting an edit.}
This repeats run 7's setting. We stopped it after 13 forward-pass calls, before it produced an edit, to run the smaller-model sessions below.

\subsection{Smaller Qwen models in the full-weight setting}

\paragraph{Run 8 edits Qwen3.5-0.8B.}
This is the original report's model, with 468 editable tensors.
For one attention head in layer 15, GPT-6 Astra identifies the 32 directions that differ most between real and black frames. It doubles the head's output component outside these directions, lengthening answers that end too early.
Its held-out practice gain does not transfer: the test difference is not significant.

\paragraph{Run 9 edits digit-related directions in Qwen3-VL-2B.}
GPT-6 Astra can edit 620 of Qwen3-VL-2B's 625 tensors. It finds wrong timestamps despite attention reaching the relevant video region.
In layer 25's MLP output projection, it selects 128 of 6,144 columns with the largest components along the nine output-layer directions that distinguish the ten digits, then subtracts 1.5 times these components.
It gains $+3.34 \pm 1.30$ on 240 development questions. GPT-6 Astra calls its held-out result, from 113 videos, inconclusive. The test difference is not significant, with $-0.23$ on VUE-TR and $-0.03$ on VUE-TR-V2.

\paragraph{Run 12 edits attention outputs in Qwen3-VL-2B.}
This independent repetition investigates wrong event locations under run 9's setting.
Single-head interventions, strength tests, sign reversal and an orthogonal-direction control identify a harmful output direction in head 13 of layer 20.
GPT-6 Astra reverses this component with $W_h' = W_h - 2.5\,u u^{\top} W_h$ and multiplies head 8's output in layer 16 by 1.5.
Likelihood tests on matched answers show a lower preference for wrong answers whose times are scaled as minutes; the gain disappears with black frames.
On held-out questions from 79 unseen videos, answers without overlap fall from 98 to 80. The test gain is significant, with $+2.16$ on VUE-TR and $+2.08$ on VUE-TR-V2.
A control preserves the edit's rank and singular values in the same two tensors but randomises the singular vectors (seed 0). It performs slightly worse than the base model ($-0.46$ $[-0.89, -0.05]$), and run 12 exceeds it by $+2.57$ $[+1.83, +3.34]$.

\paragraph{Run 14 rescales two attention-head outputs.}
This independent repetition of run 9's setting doubles head 4's output in layer 19 and multiplies head 5's by four, excluding eight directions responsive to the start of the answer.
It gains $+5.54 \pm 1.24$ on 240 development questions, but GPT-6 Astra judges the held-out result inconclusive. The test gain is nevertheless significant, with $+1.56$ on VUE-TR and $+1.43$ on VUE-TR-V2.
Its control, constructed as for run 12, performs worse than the base model ($-1.48$ $[-2.27, -0.71]$), and run 14 exceeds it by $+2.97$ $[+1.97, +4.01]$.
Thus two of the three Qwen3-VL-2B sessions improve on the test sets, and both outperform a random edit of the same size.

\subsection{TimeLens-8B in the full-weight setting}

\label{sec:tl8-sessions}

\paragraph{Three sessions independently find a dropped digit.}
Runs 15, 16 and 17 give GPT-6 Astra the same 480 practice questions and access to 744 of TimeLens-8B's 750 tensors~\cite{zhang2026timelens}. The model shares Qwen3-VL-8B's architecture, and the task gives no hint about a failure.
All three identify a decimal point written one digit too early: on a long video, an event at 1,744 to 1,795 seconds can be reported as 174 to 179 seconds.
Each submits a rank-1 edit to layer 34's MLP output projection. The paragraphs below describe runs 16 and 17, then run 15, which ran before them; runs 27 and 28 repeat the setting (\cref{sec:new-sessions}).

\paragraph{Rank-1 updates along a shared output direction.}
Run 16 reads the MLP output along the output layer's direction for ``\texttt{.}'' relative to the mean digit. It subtracts 2.3 times this reading along a copy of that direction from which the digit-distinguishing directions are removed.
Run 17 lowers the output along almost the same direction ($|\cos| = 0.99$), but the MLP input directions of the two updates differ ($|\cos| = 0.07$). Its change is 3.5 times larger.
Run 16 gains $+6.15 \pm 1.42$ on 151 of its development questions and $+1.26 \pm 0.59$ with black frames. The held-out questions cover 68 videos for run 16 and 110 for run 17 (\cref{tab:sessions}).

\paragraph{The edit of run 17 as a formula.}
Run 17 changes only layer 34's MLP output projection, also called the down-projection, $W \in \mathbb{R}^{4096 \times 12288}$. Layers are numbered from zero. The rank-1 update is
\begin{equation}
  W' = W - s\, u\, (v^{\top} W), \qquad s = 8,
\end{equation}
which changes the MLP output $y$ to $y - s\,(v^{\top} y)\,u$.
The unit output direction $u$ lies along $\gamma \odot (w_{\mathrm{dot}} - \bar{w})$: $w_{\mathrm{dot}}$ is the decimal point's output-layer row, $\bar{w}$ is the mean row of the ten digits, and $\gamma$ is the final RMSNorm gain.
Moving the final residual stream along $u$ raises the decimal point's logit relative to the digits.

The unit selector $v$ determines when the edit acts. It is constructed from the base model's layer-34 attention and MLP increments during teacher forcing on practice answers.
The editor first averages the increments at the decimal points of six long-video answers whose times are about ten times too small.
It then makes this mean orthogonal to 60 protected increments: those at 24 correctly placed decimal points, and mean increments at the answer template's other tokens, the end of the answer and the first digits of both times.
Among strengths 4, 6, 8 and 10, paired mIoU gains on 240 development questions are $+6.47$, $+8.98$, $+10.01$ and $+9.48$.
The chosen strength is fixed before evaluating held-out questions.
The stored change is rank one to float32 precision; its left singular vector has $|\cos| = 1.000$ with $u$, and its Frobenius norm is 15.1\% of $W$'s. No loss or gradient is used.

\paragraph{Test gains and the times-ten rule.}
Both edits improve on the test sets (\cref{tab:sessions,fig:transfer}); \cref{fig:qual-long} shows the six largest gains on long videos.
The times-ten rule multiplies both predicted times by ten when the answer ends in the first tenth of the video.
Run 16 performs worse than this rule ($-2.42$ $[-4.37, -0.58]$), while run 17 performs better ($+2.34$ $[+0.81, +3.81]$).
Run 17 gains $+8.53$ on VUE-TR and $+7.42$ on VUE-TR-V2; run 16 gains $+3.71$ and $+2.71$, respectively.

\paragraph{Random edits with the same norm.}
For each run, the control adds $\lVert \Delta \rVert_F\, a b^{\top}$ to the same tensor, where $\Delta$ is the applied weight change and $a$, $b$ are random unit vectors.
Both controls use the same random vectors and differ only in size. The test rule, fixed before evaluation, requires each edit to outperform its control on the test sets.
At run 16's size, the control preserves 1,155 of 1,215 answers word for word and slightly lowers mIoU ($-0.31$ $[-0.65, -0.04]$).
At run 17's size, it preserves 1,145 answers and also lowers mIoU ($-0.29$ $[-0.63, -0.01]$).
Runs 16 and 17 exceed their controls by $+3.44$ $[+2.51, +4.47]$ and $+8.17$ $[+6.56, +9.79]$ overall, and by $+6.27$ $[+4.22, +8.70]$ and $+18.37$ $[+14.84, +22.17]$ on videos of at least 30 minutes.

On the A40 machine, runs 16 and 17 gain $+3.26$ $[+2.35, +4.24]$ and $+7.82$ $[+6.26, +9.43]$ over that machine's base answers.
Three additional random controls at run 17's size use seeds 0, 1 and 2 on the A40 machine. None differs significantly from the base model: $+0.12$ $[-0.12, +0.36]$, $+0.07$ $[-0.13, +0.30]$ and $+0.02$ $[-0.23, +0.27]$.
Run 17 exceeds them by $+7.70$ $[+6.13, +9.32]$, $+7.75$ $[+6.17, +9.34]$ and $+7.80$ $[+6.20, +9.40]$, respectively.

\begin{figure}[t]
\centering
\includegraphics[width=\linewidth]{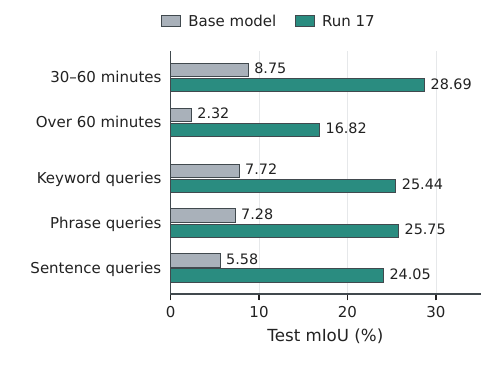}
\caption{\textbf{Run 17 improves across all official categories of long videos.} Base-model (grey) and run-17 (teal) test mIoU on 422 questions about videos of at least 30 minutes. The VUE-TR categories are long videos (30 to 60 minutes, 293 questions), ultra-long videos (over 60 minutes, 129), and keyword (111), phrase (160) and sentence queries (151). In this order, the gains are $+19.93$ $[+15.57, +24.49]$, $+14.50$ $[+8.45, +21.20]$, $+17.72$ $[+11.30, +24.12]$, $+18.47$ $[+12.69, +24.42]$ and $+18.47$ $[+13.45, +23.93]$. Across all 422 questions, the gain is $+18.27$ $[+14.74, +22.07]$, from 6.79 to 25.06 mIoU. The visual query modality is shared by all test questions and is omitted.}
\label{fig:long-video-radar}
\end{figure}

\begin{figure*}[t]
\centering
\includegraphics[width=\textwidth]{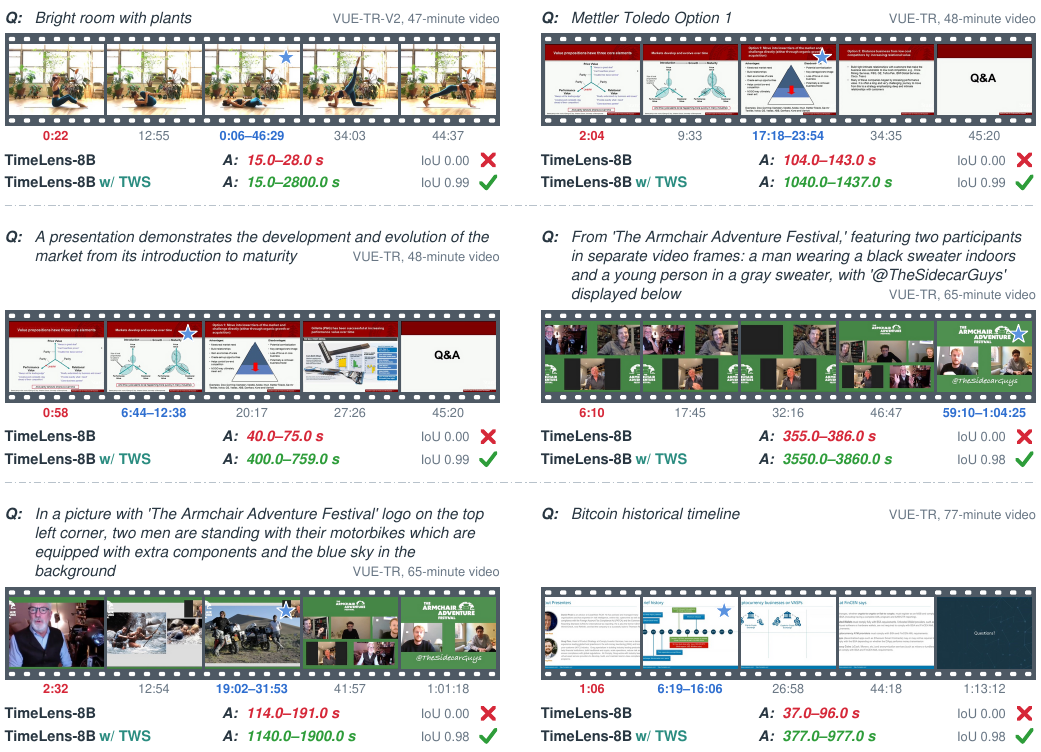}
\caption{\textbf{Run 17 restores missing digits on long videos.} The six largest IoU gains among 422 test questions about videos of at least 30 minutes, selected by rules fixed before viewing any example. Each strip shows five frames in time order. A star marks a frame within the true event, shown in blue; a red time marks the midpoint of an incorrect answer. The base and edited answers appear below exactly as written, with their IoU; a check marks IoU of at least 0.5. In all six examples, the base model omits one or two final digits of the true end time, and run 17 writes the full number.}
\label{fig:qual-long}
\end{figure*}

\paragraph{Gains by video length.}
The 1,215 test questions include 498 on videos under 10 minutes, 295 on videos of 10 to 30 minutes and 422 on videos of at least 30 minutes. In this order, run 17 gains $+0.80$ $[-0.01, +1.67]$ (not significant), $+4.99$ $[+2.28, +8.03]$ and $+18.27$ $[+14.74, +22.07]$. Run 16 gains $+0.29$ $[-0.29, +0.97]$, $+3.44$ $[+1.41, +5.54]$ and $+6.26$ $[+4.21, +8.69]$. The TimeLens-8B LoRA model of \cref{sec:lora-tl8} gains $+0.43$ $[-1.09, +1.97]$, $+5.17$ $[+2.13, +8.14]$ and $+22.90$ $[+18.92, +27.19]$.
\Cref{fig:long-video-radar} breaks the long-video gain down by benchmark category.
Main-paper Figs.~3 and~4 use a separate evaluation, where run 17 gains $+0.30$ and $+18.51$ on the shortest and longest videos. Each reported difference pairs answers from its own evaluation.

\paragraph{Short videos and MomentSeeker.}
On the 297 Charades-TimeLens questions about videos of 9 to 59 seconds, none of runs 15, 16 and 17 differs significantly from the base model (46.61 mIoU).
Their gains are $+0.31$ $[-0.57, +1.17]$, $+0.33$ $[-0.43, +1.21]$ and $-0.10$ $[-1.18, +0.96]$, with 275, 276 and 253 answers unchanged word for word, respectively.
On the 279 MomentSeeker questions about videos of at least 10 minutes, run 17's difference is also not significant ($10.09 \to 10.72$, $+0.64$ $[-1.29, +2.36]$). Runs 15 and 16 were not evaluated on MomentSeeker.

\paragraph{Run 15 protects answers that start at zero.}
Run 15 ran before runs 16 and 17, under the same task and settings, with the same model and practice questions. It also identifies the dropped digit without a hint.
Its edit changes only layer 34's MLP output projection, $W' = W - 3\,u (\tilde{v}^{\top} W)$, with a selector $\tilde{v}$ that protects answers starting at zero and formatting tokens.
A stronger version without this protection produced loops of zeros and was rejected.
The edit is fixed before confirmation on 113 unseen videos. Answers off by a factor of ten fall from 30 to 6; with black frames it gains nothing.
The output direction aligns with the output-layer direction for the decimal point relative to digits ($|\cos| = 1.00$, as in run 17).
Its selector has $|\cos| = 0.73$ with the output direction, and its relative edit norm, 0.088 of the tensor's norm, lies between those of runs 16 and 17.

Run 15 gains $+5.43$ on VUE-TR and $+5.75$ on VUE-TR-V2.
On videos of at least 30 minutes, mIoU rises from 6.79 to 21.24 ($+14.44$ $[+11.48, +17.81]$).
The difference from the times-ten rule is not significant overall ($+0.07$ $[-1.68, +1.72]$), but is negative on videos of at least 30 minutes ($-6.59$ $[-10.38, -2.86]$).
A random rank-1 edit of run 15's size, constructed as above for runs 16 and 17, does not differ significantly from the base model ($+0.12$ $[-0.17, +0.44]$).
Run 15 exceeds this control by $+5.49$ $[+4.22, +6.87]$.

\paragraph{Unconditional rescaling and a times-ten oracle.}
Multiplying every predicted time by ten gives 19.07 mIoU, below runs 16 and 17 by $13.61$ and $18.37$.
An oracle chooses between each original prediction and its times-ten version using the true answer. This upper bound, unavailable in practice, scores 40.54 and exceeds runs 16 and 17 by $7.86$ and $3.11$. It chooses only between the answer and its times-ten version; the scale ceiling of \cref{sec:headroom} ($+12.0$) also tries times 60, $1/10$ and $1/60$.
On videos of at least 30 minutes, run 17 does not differ significantly from unconditional multiplication by ten ($-2.25$ $[-5.32, +0.76]$) or the times-ten rule applied only when the answer ends in the first tenth of the video ($-2.87$ $[-5.82, +0.01]$). Neither rule uses the true test answer.

\subsection{Screening of further models}
\label{sec:screening}

\paragraph{Screening for correctable errors.}
Before editing a new model, we evaluate it on 120 practice questions: 60 about videos of 10 to 30 minutes and 60 about longer videos. We use 448 frames, greedy decoding and the standard prompt (\cref{sec:compute}).
The rule, fixed before each screening, starts a GPT-6 Astra session only when an estimated repair could gain at least 5 mIoU.
We take the better of two upper bounds: correcting the timestamps (keeping them as written, multiplying by ten, reading them as minutes, or using one scale factor for all answers), and, for thinking models, completing answers cut at the token cap.
These estimates use the true answers and are not deployable methods.

Qwen2.5-VL-7B-Instruct~\cite{bai2025qwen25vl} encodes frame times in position IDs rather than text timestamps. It scores 3.54 mIoU, against 21.25 for Qwen3-VL-8B-Instruct on the same questions.
Its answers end near two minutes regardless of video length (median 104 seconds on videos of at least 30 minutes), and timestamp correction gains at most 3.2.
GLM-4.1V-9B-Thinking~\cite{hong2025glm41v}, whose language model is outside the Qwen family, scores 22.68, with 4 answers cut at 2,048 tokens. Timestamp correction gains at most 2.65.
Neither receives an editing session: Qwen2.5-VL-7B-Instruct rarely finds the event, while GLM-4.1V-9B-Thinking leaves little room for an answer-format repair.

Other Qwen models also remain below the threshold.
Qwen3-VL-2B-Thinking scores 17.10, with 13 answers cut at 2,048 tokens and a maximum estimated repair gain of 2.08.
Qwen3-VL-4B-Thinking scores 26.08, with 7 answers cut and a maximum gain of 1.62; Qwen3-VL-4B-Instruct scores 17.10, with a maximum gain of 2.70.
Qwen3-VL-8B-Thinking scores 24.60, with 4 answers cut at 2,048 tokens and a maximum gain of 0.85.

Two additional families remain below the threshold.
MiMo-VL-7B~\cite{xiaomi2025mimovl} (2508 release~\cite{xiaomi2025mimovl2508}) uses its own language model and reasons before answering. The reinforcement-learning version scores 12.00, with 7 answers cut; one scale factor for all answers gains at most 3.16. The supervised version scores 7.31, with a maximum gain of 3.78.
InternVL3.5-8B~\cite{wang2025internvl35} uses its own vision encoder and a Qwen3-8B language model. For this model, we use 64 frames rather than 448, supplying each as an image labelled with its time in our prompt. It scores 15.89, with a maximum gain of 1.42. In thinking mode these values are 18.49 and 0.84.
TimeLens-8B, by comparison, scores 29.24 on these questions, with an estimated repair gain of 12.39, above the screening threshold.

\paragraph{Editing models below the screening threshold.}
To test whether a correctable failure is needed, we also run GPT-6 Astra in the full-weight setting on InternVL3.5-8B (run 21) and MiMo-VL-7B-SFT-2508 (run 22).
Run 22 allows up to 1,024 tokens and reads the span after the reasoning. Both tasks name no failure, and test rules are fixed before the sessions.
Run 21 identifies 32 directions from the effect of ablating one layer-23 head on answer digits, then removes these directions from the head's output.
Its held-out practice gain does not transfer: the test difference is not significant, with $-0.03$ on VUE-TR and $+0.21$ on VUE-TR-V2.
A control edits the same tensor and head, preserving the edit's singular values but randomising its directions (seed 0). Neither its difference from the base model ($-0.04$ $[-0.33, +0.21]$) nor run 21's difference from it ($+0.14$ $[-0.26, +0.51]$) is significant.
Run 22 stops before submitting an edit.
All 26 candidates evaluated on its 162 development questions perform at or below the base model: changes range from $-3.20$ to $0.00$ mIoU against a base score of 11.01.

\subsection{Qwen3.5-0.8B in thinking mode}

\paragraph{Repeated reasoning and decoding baselines.}
With thinking mode and a limit of 2,048 new tokens, 48 of Qwen3.5-0.8B's 120 screening answers reach the limit before finishing their reasoning (\cref{tab:thinking}).
As in the weight-space programming report, these answers repeat themselves until the token limit; here they list timestamps: ``At 22:45, Grandpa Pig is holding a stick. At 22:50, Grandpa Pig is holding a stick. \ldots''.
Of the 72 completed answers, 71 finish within 1,024 tokens.
If truncated answers finished with the same accuracy as completed ones, the gain would be 10.02, above the screening threshold.

We compare thinking mode with two decoding baselines, non-thinking mode and budget forcing~\cite{muennighoff2025s1}, which closes reasoning still open at 1,024 tokens and lets the model answer. On practice questions, budget forcing gains $+2.92 \pm 1.24$ over thinking mode; the forced answers contain a span but score only 7.15.

On the test sets, with a limit of 1,024 new tokens, 447 of 1,215 answers (37\%) reach the limit. Non-thinking mode performs worse than thinking mode, while budget forcing improves it (\cref{tab:thinking,fig:reasoning}). Budget forcing preserves word for word the 768 answers that finish on their own.

\begin{table*}[t]
\centering
\caption{\textbf{Decoding baselines for repeated reasoning in Qwen3.5-0.8B.} mIoU on the 120 screening questions (practice) and on the 1,215 test questions. ``Hit token limit'' counts answers whose reasoning is still open at the limit (2,048 tokens on practice in thinking mode, 1,024 otherwise). Test $\Delta$mIoU is paired against thinking mode, with 95\% bootstrap confidence intervals over videos.}
\label{tab:thinking}
\small
\setlength{\tabcolsep}{3pt}
\begin{tabular}{lcccc}
\toprule
 & \multicolumn{2}{c}{Practice (120)} & \multicolumn{2}{c}{Test (1,215)} \\
Decoding method & mIoU & Hit token limit & mIoU & $\Delta$mIoU \\
\midrule
Thinking mode & 15.03 & 48 & 18.67 & \\
\quad answers within the limit & 25.04 & & 29.54 & \\
Non-thinking mode & 12.84 & -- & 15.36 & $-3.31$ $[-5.21, -1.38]$ \\
Budget forcing & 17.95 & 0 & 21.27 & $+2.59$ $[+1.97, +3.25]$ \\
Repetition penalty 1.1 & 17.89 & 32 & 19.59 & $+0.92$ $[-0.81, +2.67]$ \\
Run 18's edit & -- & -- & 21.00 & $+2.33$ $[+0.88, +3.79]$ \\
Run 19's edit & -- & -- & 19.92 & $+1.24$ $[-0.14, +2.70]$ \\
\bottomrule
\end{tabular}
\end{table*}

\paragraph{Setting of runs 18 and 19.}
Both sessions give GPT-6 Astra full-weight access to Qwen3.5-0.8B in thinking mode, the same 480 practice questions and no hint about repeated reasoning.
Answers allow at most 1,024 new tokens, and the span is read after the reasoning.
In thinking mode, generation calls contain at most 64 questions, and each session has 30 calls.
Rules fixed before the sessions compare each submitted edit on the test sets against thinking mode, non-thinking mode and budget forcing.

\paragraph{Run 18 removes a direction associated with repetition.}
Half of the session's diagnostic answers run out of tokens while repeating their reasoning.
Removing whole heads associated with copying helps answers finish but does not improve grounding. GPT-6 Astra therefore removes a single direction $u$, estimated from layer-20 residual differences between repeated phrases and their first occurrences.
It applies $W' = W - u(u^{\top} W)$ to the attention output projections of layers 16 to 23, comprising six Gated DeltaNet layers and two full-attention layers.
The weight-space programming report~\cite{takehi2026sparks} removes a repetition direction from 64 MLP output columns in each of layers 15 and 19 in its session with 50 forward-pass calls, and from attention and MLP outputs in layers 8 to 22 in its session with 10 calls.
Here, the edit covers all attention output columns in layers 16 to 23. Editing MLP outputs alone or using a random direction of the same norm did not help during the session.

With the edit fixed, truncated answers on 131 held-out videos fall from 106 to 65. Run 18 also improves on the test sets (\cref{tab:thinking}), with 863 rather than 768 answers finishing their reasoning.
It also exceeds non-thinking mode ($+5.64$ $[+3.61, +7.60]$), but does not differ significantly from budget forcing ($-0.27$ $[-1.82, +1.26]$), which needs no editor.

A repetition penalty of 1.1 is chosen over 1.2 on the screening questions (17.89 versus 12.24 mIoU).
This standard decoding control lets 870 test answers finish, a similar number to the edit, but its completed answers score 27.36, against 29.54 for thinking mode's completed answers.
Neither its difference from thinking mode nor run 18's advantage over it ($+1.41$ $[-0.36, +3.15]$) is significant.
A random edit preserves the rank and singular value in each of the eight tensors but randomises the directions (seed 0).
It yields 744 completed answers and no significant difference from thinking mode ($+0.26$ $[-1.01, +1.50]$); run 18 exceeds it by $+2.06$ $[+0.52, +3.63]$, which is not significant after Holm correction (\cref{sec:holm}).

\paragraph{Run 19 halves the output of one head.}
Under the same rules, this session halves the output of head 7 in layer 15. Attention probes link this head to copying repeated reasoning; the report's session with 25 calls zeroed the same head~\cite{takehi2026sparks}.
Removing it entirely hurts grounding during the session.
GPT-6 Astra reports no reliable held-out improvement, and the test difference from thinking mode is also not significant (\cref{tab:sessions,tab:thinking}). On the test sets, 801 answers finish. The edit exceeds non-thinking mode ($+4.55$ $[+2.58, +6.64]$) but does not differ significantly from budget forcing ($-1.35$ $[-2.85, +0.19]$).

\section{Comparisons across editing sessions}
\label{sec:visual-comparisons}

Among the 16 tested sessions that edit base weights (\cref{tab:sessions}), 2 change query projections, 10 change value or attention-output projections (run 4 changes the value and output projections), and 4 change MLP output projections. Run 1 edits an adapter and is not counted.

\begin{figure*}[t]
\centering
\includegraphics[width=\textwidth]{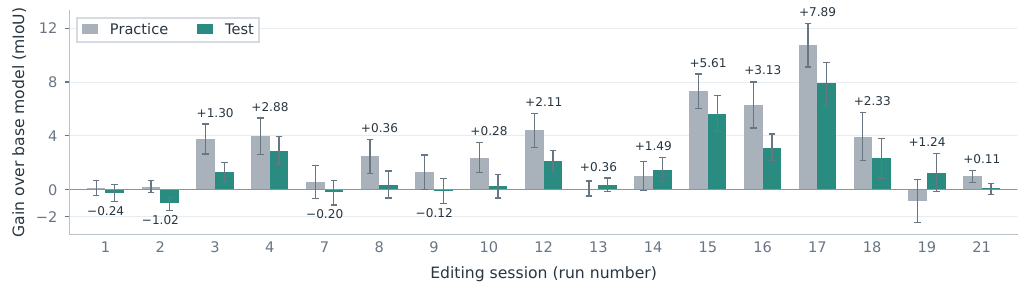}
\caption{\textbf{Practice gains do not always transfer to unseen videos.} The 17 tested sessions in \cref{tab:sessions}, in run-number order. Grey bars show reported practice gains with one standard error; teal bars show paired test gains over the base model with 95\% bootstrap intervals over videos. Labels give the test gains. A test interval above zero indicates improvement, one below zero degradation, and one containing zero no significant difference. Run 1 used 478 practice questions with none held out, base answers were inspected before run 2, and run 4 reused practice questions to select strength. Because models and practice subsets differ across sessions (\cref{tab:sessions}), the figure does not provide a controlled comparison of generalisation across sessions.}
\label{fig:practice-test}
\end{figure*}

Runs 15 to 17, the TimeLens-8B MLP edits, have the largest gains among these sessions. This does not isolate the causal effect of choosing a projection or layer. Practice and test gains (\cref{fig:practice-test}) are not directly comparable across different models and practice subsets. The main paper's statement that the highest practice gain selects the best test edit refers to Qwen3-VL-8B, Qwen3-VL-2B, TimeLens-8B and Qwen3.5-0.8B; runs 1 and 2 on TimeLens2-8B start from different weights and are not counted.

\Cref{tab:session-means} gives each model's mean and best test gain across tested sessions.
The five TimeLens-8B sessions, including runs 27 and 28 of \cref{sec:new-sessions}, average $+5.45$, close to the times-ten rule's $+5.55$. Individually, runs 17 and 27 outperform the rule ($+2.34$, $+1.75$), run 15 shows no significant difference ($+0.07$), and runs 16 and 28 perform worse ($-2.42$, $-2.25$).

\begin{table}[t]
\centering
\caption{\textbf{Mean and best test gains across every tested session for each model.} Each gain uses all 1,215 questions and pairs an edit with the base model on the same machine (\cref{tab:sessions,tab:new-sessions}). Sessions without an edit are excluded. The TimeLens2-8B row includes run 1's adapter edit.}
\label{tab:session-means}
\small
\setlength{\tabcolsep}{3pt}
\resizebox{\columnwidth}{!}{%
\begin{tabular}{lrrr}
\toprule
Base model and editor & Sessions & \shortstack{Mean\\$\Delta$mIoU} & \shortstack{Best\\$\Delta$mIoU} \\
\midrule
TimeLens-8B & 5 & $+5.45$ & $+7.89$ \\
Qwen3-VL-8B with Claude Opus 5.5 & 2 & $+2.09$ & $+2.88$ \\
Qwen3-VL-8B with GPT-6 Astra & 3 & $+0.15$ & $+0.36$ \\
Qwen3-VL-2B & 3 & $+1.16$ & $+2.11$ \\
Qwen3.5-0.8B in thinking mode & 2 & $+1.79$ & $+2.33$ \\
Qwen3.5-0.8B in non-thinking mode & 1 & $+0.36$ & $+0.36$ \\
InternVL3.5-8B & 1 & $+0.11$ & $+0.11$ \\
TimeLens2-8B & 2 & $-0.63$ & $-0.24$ \\
\bottomrule
\end{tabular}}
\end{table}

\section{Where grounding errors remain}
\label{sec:headroom}

Unless stated otherwise, this section uses answers saved on the A40 machine. We measure the room for further timestamp repairs and test whether attention locates events that generated answers miss.
The oracle repairs use the true intervals, so they measure room for improvement and cannot be applied at test time.

\subsection{A census of the errors of the base models}

\Cref{tab:census} summarises errors on the 1,215 test questions. An answer \emph{collapses to the start} if it ends in the first 2\% of the video. We report four oracle gains, called ceilings in the table. The \emph{centre} column gives the gain from moving each predicted interval to the true centre while retaining its length. The \emph{length} ceiling replaces its length with the true length while retaining its centre. The \emph{scale} ceiling selects the best of the original answer and its times multiplied by 10, 60, $1/10$ or $1/60$. The \emph{frame-index} ceiling selects the better of the original answer and its times interpreted as indices of the 448 frames.
Scaling the length of every predicted interval about its centre by one common factor, even the best factor from 0.25 to 4, improves no model.

TimeLens-8B's scale ceiling is $+12.0$ overall, the perfect repair of main-paper Sec.~5.1, and $+26.3$ on videos of 30 minutes or more. Collapsed Qwen3-VL-8B and Qwen3-VL-2B answers have median ratios of true to written start times of 52.5 and 50, consistent with minutes written as seconds.
Run 17 raises TimeLens-8B to 37.36 mIoU. The remaining oracle gains are $+3.4$ from correcting the scale and $+11.4$ from correcting the interval centre; 24.4\% of answers still miss the event.
Correcting the interval centre offers every base model the largest gain; 40 to 47\% of non-thinking answers miss the event.

\begin{table*}[t]
\centering
\caption{\textbf{Error census of the base models on the 1,215 test questions.} Ceilings are mIoU gains of oracle repairs that use the true interval, averaged over all questions. In thinking mode, 37.8\% of Qwen3.5-0.8B's answers give no span. No overlap counts answers with a span that misses the event. The Qwen3.5-0.8B rows use the L40S answers of \cref{tab:thinking}.}
\label{tab:census}
\small
\setlength{\tabcolsep}{5pt}
\begin{tabular}{lccccccc}
\toprule
 & & \multicolumn{2}{c}{Answers} & \multicolumn{4}{c}{Ceiling of a repair} \\
Base model & mIoU & \shortstack{collapse\\to start} & \shortstack{no\\overlap} & centre & length & scale & \shortstack{frame\\index} \\
\midrule
Qwen3-VL-8B & 24.62 & 27.2\% & 43.1\% & $+20.8$ & $+6.5$ & $+5.4$ & $+1.4$ \\
TimeLens-8B & 29.54 & 14.7\% & 40.8\% & $+16.8$ & $+4.4$ & $+12.0$ & $+2.7$ \\
Qwen3-VL-2B & 21.30 & 26.0\% & 42.5\% & $+19.2$ & $+7.3$ & $+6.9$ & $+1.1$ \\
InternVL3.5-8B & 21.90 & 8.8\% & 44.4\% & $+18.3$ & $+7.5$ & $+3.3$ & $+1.1$ \\
Qwen3.5-0.8B, non-thinking & 15.36 & 14.7\% & 47.2\% & $+21.4$ & $+10.3$ & $+4.7$ & $+0.7$ \\
\quad thinking mode & 18.67 & 6.3\% & 20.8\% & $+10.8$ & $+3.7$ & $+2.9$ & $+1.3$ \\
\bottomrule
\end{tabular}
\end{table*}

\begin{figure}[t]
\centering
\includegraphics[width=\linewidth]{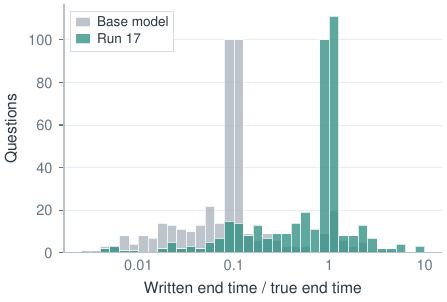}
\caption{\textbf{Run 17 shifts end times towards the correct scale.} Written end time divided by true end time, on a logarithmic axis, for the 422 test questions on videos of 30 minutes or more. Within a factor of 1.4 of each value (0.15 on the log axis), 51.4\% of base answers lie near one tenth of the true end time and 7.3\% near the true end time. After the edit, these shares are 8.5\% and 53.3\%.}
\label{fig:end-time-ratio}
\end{figure}

\paragraph{Errors by event position in long videos.}
On videos of 30 minutes or more, TimeLens-8B scores 14.8 mIoU on events in the first fifth, 3.0 in the middle three fifths and 0.0 in the last fifth, where no answer overlaps the event. Run 17 raises these to 24.8, 26.1 and 23.5, and \cref{fig:end-time-ratio} shows how it moves end times towards the correct scale.
The base model thus performs worst at the end, unlike the U-shaped bias of long-context language models, which often perform worst in the middle~\cite{liu2024lost}.
On the same long videos, Qwen3-VL-8B scores about 10 mIoU with or without steering: 40\% of its long-video answers end in the first 2\%, and 68 of these 169 start at zero, as in ``\texttt{0.0 - 10.0 seconds}''.

\paragraph{Few events fall between the sampled frames.}
For 72 of the 1,215 test events (5.9\%), and for 56 of the 422 on videos of 30 minutes or more (13.3\%), none of the 448 frames lies inside the true interval, and the models score 1 to 2 mIoU there.
Most missed events are therefore shown to the model.

\paragraph{Events that cover half the video or more.}
In 189 test questions (15.6\%) the true event covers half the video or more; 63\% of these questions are on videos under 10 minutes, and 97 of the queries are keywords, such as ``\texttt{blue lighting}'', for something visible throughout.
Predicting the whole video scores 91.0 mIoU on these questions, compared with 20.6 for Qwen3-VL-2B, 38.8 for Qwen3-VL-8B, 51.2 for TimeLens-8B and 66.0 after run 17's edit. Qwen3-VL-2B predicts less than a tenth of the video for 69\% of them.
Keeping, on these questions, the better of each answer and the whole video would add $+11.0$, $+8.2$, $+6.4$ and $+4.2$ mIoU overall, respectively.
Video length alone does not identify these questions: predicting the whole video for every video under a minute loses 0.97 to 2.82 mIoU, because most events in short videos are still short.
In saved answers of the TimeLens2 evaluation~\cite{zhu2026timelens2} at 448 frames on the TimeLens versions of ActivityNet and QVHighlights~\cite{zhang2026timelens}, Qwen3-VL-8B scores 66.5 and 64.8 on events covering half the clip or more, so long events alone do not explain these errors.
Every practice video lasts at least 10 minutes. Only 2 of the 480 practice questions (0.4\%) have such an event, so the sessions had almost no practice examples of this failure.

\paragraph{Why the repairs do not improve the other datasets.}
Charades-TimeLens and MomentSeeker offer little room for these timestamp repairs (main-paper Sec.~5.4).
The 297 Charades-TimeLens questions are not a random sample: they make up 156 pairs with the same object and the opposite action (76 open and close, 71 take and put, 6 turn on and turn off, 3 sit and stand), some questions belonging to two pairs, across 133 clips lasting 9 to 59 seconds.
No timestamp needs minutes, and Qwen3-VL-8B never uses m:ss with or without an edit. No TimeLens-8B answer ends in the first 2\% of the clip; its scale ceiling is $+2.6$.
MomentSeeker's 279 questions cover 31 videos of 10 to 100 minutes (median 38 minutes per question). The edits change answers as intended: Qwen3-VL-8B's m:ss answers rise from 152 to 204 with the query edit and 239 with the value/output edit. Run 17 reduces TimeLens-8B answers ending in the first 2\% from 35.5\% to 8.2\%.
Yet the base models score only 7.26 and 10.09 mIoU. For TimeLens-8B, 74.6\% of answers miss the event, with a scale ceiling of $+2.1$ versus $+38.5$ from correcting the interval centre.
The query and value/output edits score 7.76 and 8.02; edited TimeLens-8B scores 10.72. Most answers still miss the event. The edits do not significantly improve either dataset.

\subsection{Changing only the separator}
\label{sec:supp-separator}

The probe in main-paper Sec.~4.2 detects m:ss when Qwen3-VL-8B writes minutes as seconds. We test whether the model also holds the correct time by changing one token, with weights fixed.
Using the same frames, prompt, bfloat16 weights and greedy decoding reproduces all 1,695 practice and test answers word for word.
After the start time's integer digits, we record the log-probabilities of the chosen token and the colon. If another token was chosen, we cut the answer at that point, insert ``\texttt{:}'' and continue greedily to produce a \emph{separator-only answer}.
A rule fixed before seeing results selects this answer when the colon's log-probability is within $b$ of the chosen token's, equivalent to a bias $b$ at that position. With $b = 15$, this is the close separator choice of main-paper Fig.~5a.
On 480 practice questions, mIoU rises from 21.55 to 23.58 at $b=6$ and 23.84 at the selected $b=15$. Always inserting a colon gives 19.69.

The rule changes 294 of 1,215 test answers, including 143 on videos of 30 minutes or more (\cref{tab:separator}). It gains $+1.47$ $[+0.65, +2.33]$ overall, but not significantly on long videos ($+0.61$ $[-0.50, +1.82]$).
Neither the query edit nor steering differs significantly from it ($+0.44$ $[-0.62, +1.49]$ and $+0.60$ $[-0.65, +1.85]$ in their favour). The value/output edit exceeds it by $+1.38$ $[+0.44, +2.36]$.
On changed answers, separator-only generation exceeds reading the original times as minutes by $+9.44$ $[+5.12, +13.91]$, so correcting the separator recovers more accurate timestamps.
The first number already matches the true start minute in 131 cases (45\%) and lies within one minute in 159 (54\%); random pairing among these answers gives 47.8 (16\%) and 59.7 (20\%).
Separator-only answers exactly match the query edit in 45 of 294 cases and the value/output edit in 108.

\begin{table}[t]
\centering
\caption{\textbf{Changing only the separator.} Qwen3-VL-8B test mIoU, all measured on the A40 machine. Columns show all 1,215 questions and the 294 whose answers the decoding rule changes.}
\label{tab:separator}
\footnotesize
\setlength{\tabcolsep}{4pt}
\begin{tabular}{lcc}
\toprule
 & All 1,215 & Changed 294 \\
\midrule
Base model & 24.62 & 13.10 \\
Base answers read as minutes & -- & 9.75 \\
Separator only (decoding rule, $b=15$) & 26.09 & 19.20 \\
Query edit & 26.53 & 16.52 \\
Value/output edit & 27.47 & 18.84 \\
Steering & 26.69 & 14.68 \\
\bottomrule
\end{tabular}
\end{table}

\paragraph{Changing only the decimal point in TimeLens-8B.}
With weights fixed, we replace a decimal point after the integer digits with the most likely digit and continue greedily, first at the start time and then at the end generated on that path.
The rule replaces it when its log-probability exceeds the digit's by less than $b$. The rule was fixed before seeing results; all 1,695 base answers match the saved ones exactly.
On 480 practice questions, mIoU rises from 30.32 to 38.88 at the selected $b=10$; always replacing the decimal point gives 1.45.

The rule changes 532 test answers and gains $+6.35$ $[+4.74, +7.96]$ overall and $+17.53$ $[+13.79, +21.37]$ on videos of 30 minutes or more. It does not differ significantly from the times-ten rule ($+0.80$ $[-0.55, +2.13]$) or the rule written from answers alone (\cref{sec:compute}; $+0.31$ $[-1.07, +1.72]$).
Run 17 exceeds it by $+1.47$ $[+0.61, +2.38]$ overall, but not significantly on long videos ($+0.98$ $[-0.73, +2.83]$). Run 15 also shows no significant difference; run 16 ($-3.09$ $[-4.53, -1.69]$) and steering ($-1.55$ $[-3.00, -0.13]$) perform worse.
On the 532 changed answers, the rule scores 21.24 mIoU, compared with 6.74 for base answers, 23.81 after multiplying their times by ten, and 24.48 for run 17. However, of the 428 changed answers whose start gains a digit, only 6 reach the true start, and 38 come within 2 seconds of it.
Unlike Qwen3-VL-8B's first number, the added TimeLens-8B digit seldom recovers the exact time. Adding a digit, rescaling the timestamps or editing the weights can correct the scale error, but these results do not establish reliable recovery of the missing digit.

\subsection{Other models under the TimeLens2 evaluation}

\Cref{tab:protocol-census} uses our saved answers from the official TimeLens2 evaluation~\cite{zhu2026timelens2}, with its frame cap lowered from 2,048 to 448 and its other settings kept: at most one frame per second, frame area at most $480 \times 480$ pixels, and answers as span lists.
That evaluation uses its own visual subsets of VUE-TR and VUE-TR-V2, 525 and 715 questions, rather than the 562 and 727 of \cref{sec:compute}; we score the 1,174 of them (491 and 683) whose videos we could download. These scores describe this different protocol, rather than our test setting.
The TimeLens2 models' scale ceilings are only $+1.7$ to $+2.3$. Qwen3-VL-8B collapses to the start more often than in our setting (48.7\% against 27.2\%).
Some models repeatedly list spans until the token limit~\cite{holtzman2020degeneration}: 21.6\% of Qwen3-VL-4B and 8.2\% of Qwen3-VL-8B answers never close their lists. Keeping the spans already written would add only $+1.51$ and $+0.65$.

\begin{table}[t]
\centering
\caption{\textbf{Error census under the TimeLens2 evaluation.} The frame cap is 448. Saved answers on 1,174 visual VUE-TR and VUE-TR-V2 questions. ``No span'' counts answers without a readable span, mostly span lists cut at the token limit. Ceilings as in \cref{tab:census}.}
\label{tab:protocol-census}
\footnotesize
\setlength{\tabcolsep}{3pt}
\begin{tabular}{lccccc}
\toprule
Model & mIoU & \shortstack{Collapse\\to start} & No span & \shortstack{Scale\\ceiling} & \shortstack{Centre\\correction} \\
\midrule
TimeLens2-2B & 41.17 & 6.2\% & 0.5\% & $+2.33$ & $+11.53$ \\
TimeLens2-4B & 43.71 & 4.5\% & 0.7\% & $+1.73$ & $+11.17$ \\
TimeLens2-8B & 44.13 & 5.2\% & 0.0\% & $+1.90$ & $+10.72$ \\
Qwen3-VL-4B & 21.23 & 4.3\% & 30.7\% & $+1.91$ & $+11.62$ \\
Qwen3-VL-8B & 16.00 & 48.7\% & 12.9\% & $+9.42$ & $+19.89$ \\
Qwen3-VL-30B-A3B & 31.05 & 12.5\% & 14.1\% & $+2.39$ & $+15.44$ \\
\bottomrule
\end{tabular}
\end{table}

\subsection{Attention to the events the models miss}

Du~\etal~\cite{du2026know} identify heads that attend from the query to the true interval during the prompt pass even when the answer is wrong. We test their published Qwen3-VL-8B, TimeLens-8B and MiMo-VL-7B heads on 422 visual test questions about videos of 30 minutes or more.
They leave the Qwen3-VL-8B and MiMo-VL variants unspecified. We use Qwen3-VL-8B-Instruct and MiMo-VL-7B-RL-2508~\cite{xiaomi2025mimovl2508}, which supports their prompt option for turning reasoning off.
Without released code, we reimplement the attention step of their method: two frames per second, at most 14,336 visual tokens, their prompt, greedy decoding and each head's positive attention log-ratio between real and all-zero frames (their Eq.~6).
Before seeing results, we set thresholds for answers with no overlap: attention peaks inside a true span in at least 30\% would warrant further study; below 15\% would end the investigation.
We also identify misses that remain after all four rescalings used for the scale ceiling above: none of the rescaled predicted intervals overlaps a true span.

\begin{table*}[t]
\centering
\caption{\textbf{Attention peaks inside events that the generated answers miss.} Top: the share of answers with no overlap whose attention peak lies inside a true span, with question counts in parentheses. Groups use the event's share of video duration, which also defines chance. Bottom: mIoU of written answers and of intervals read from the same attention with Du~\etal's interval read-out. Each model is evaluated on all 422 test questions about videos of 30 minutes or more, under Du~\etal's input setting (two frames per second, at most 14,336 visual tokens) rather than our 448-frame test setting. MiMo-VL-7B is MiMo-VL-7B-RL-2508.}
\label{tab:attention}
\small
\setlength{\tabcolsep}{6pt}
\begin{tabular}{lccc}
\toprule
 & Qwen3-VL-8B & TimeLens-8B & MiMo-VL-7B \\
\midrule
\multicolumn{4}{l}{\emph{Peak inside a true span, answers without overlap}} \\
\quad all & 44.6\% (271) & 41.8\% (294) & 55.5\% (373) \\
\quad event under 2\% of the video (chance 0.7\%) & 34.6\% (182) & 27.9\% (190) & 44.3\% (237) \\
\quad event 2 to 10\% of the video (chance 5\%) & 65.5\% (58) & 68.4\% (76) & 77.8\% (99) \\
\quad no overlap after any tested rescaling & 40.1\% (187) & 11.9\% (118) & 52.5\% (282) \\
\midrule
\multicolumn{4}{l}{\emph{mIoU on all questions}} \\
\quad written answer & 12.33 & 12.29 & 1.29 \\
\quad interval read from the attention & 14.61 & 12.82 & 14.44 \\
\bottomrule
\end{tabular}
\end{table*}

Every model exceeds the 30\% threshold (\cref{tab:attention}). Each also exceeds chance for events covering under 2\% of the video. For TimeLens-8B answers that still miss the event after every tested rescaling, however, only 11.9\% have an attention peak inside it.
Decoding intervals from attention requires extracting attention scores during inference; the edited model instead uses its usual decoding procedure. Decoding from attention gains $+2.28$ for Qwen3-VL-8B, $+0.53$ for TimeLens-8B and $+13.15$ for MiMo-VL-7B, whose answers are often ``\texttt{-1 - 0 seconds}''.
Du~\etal\ report the opposite pattern for Qwen3-VL-8B on short videos: 51.7 from attention versus 59.4 from generated answers on QVHighlights-TimeLens~\cite{du2026know}. On long videos, attention often peaks inside events that the written timestamps miss.

\paragraph{Scaling the heads by a weight edit.}
We next test whether scaling the published heads makes the generated time follow their attention. For TimeLens-8B and Qwen3-VL-8B, we multiply each head's output by $g$ through its 128 columns in the attention output projection.
As for steering, we select $g \in \{0.5, 1.5, 2, 3\}$ on development questions, check it on held-out questions and evaluate it once on the test sets. If test performance improves, we compare with the same scaling applied to the same number of other heads (five) in the same layers. These rules were fixed before seeing results.

Amplification worsens development accuracy. At $g = 1.5$, 2 and 3, changes are $-3.68$, $-9.09$ and $-21.55$ for TimeLens-8B, and $-3.25$, $-6.15$ and $-16.26$ for Qwen3-VL-8B.
Halving gives $+1.37 \pm 0.76$ and $-2.22 \pm 1.01$, respectively. Following the predefined rule, we evaluate $g=0.5$ for both models because it gives the highest development score, even though all candidates reduce Qwen3-VL-8B's development accuracy. Held-out changes are $-1.42 \pm 0.94$ and $-2.25 \pm 1.19$; TimeLens-8B's positive development gain is not reproduced on the held-out questions.
On the test sets, halving TimeLens-8B's published heads gains $+0.95$ $[+0.02, +1.88]$, but halving five other heads in the same layers (seed 0) also gains $+0.73$ $[+0.10, +1.39]$. Their difference is not significant ($+0.22$ $[-0.59, +1.02]$), so this comparison does not establish an advantage specific to the published heads. Qwen3-VL-8B shows no significant change ($+0.65$ $[-0.24, +1.51]$).

\section{Additional experiments and limits of the repairs}
\label{sec:evidence}

We fixed the rules for each experiment before examining its results and report every condition we ran.
Unless stated otherwise, gains use the main paper's metric on the 1,215 test questions, with 95\% bootstrap intervals over videos. Captions name the hardware, and comparisons always pair answers from the same GPUs.

\subsection{Training timestamp range and dropped digits}
\label{sec:why-digit}

TimeLens-8B was trained on TimeLens-100K at 2 frames per second, with at most 448 frames and videos of at most 500 seconds~\cite{zhang2026timelens}.
None of the 193,172 target times reaches 1,000 seconds; the largest is 462 seconds (\cref{tab:digits}). This training set therefore contains no times with four integer digits.
In the base model's test answers, 273 of the 489 written times whose true time is at least 1,000 seconds (55.8\%) have three integer digits followed by a decimal point, with the true time within 15\% of ten times the predicted integer part.
These counts link the dropped digit to times outside the training range: the decimal point ends the integer part at the longest length seen in training.

On the 422 questions about videos of at least 30 minutes, four-digit times account for 70.9\% of true end times, 2.1\% of the base model's predicted end times, 10.0\% of run 16's, 51.2\% of run 17's and 56.4\% of the TimeLens-8B LoRA model's (\cref{sec:lora-tl8}). One run 16 answer has no readable time.

\begin{table}[t]
\centering
\caption{\textbf{TimeLens-100K contains no four-digit target times.} Top: training targets by number of integer digits. Bottom: base test timestamps with three integer digits followed by a decimal point, for which the true time is at least 1,000 seconds and within 15\% of ten times the predicted integer part.}
\label{tab:digits}
\small
\setlength{\tabcolsep}{4pt}
\resizebox{\columnwidth}{!}{%
\begin{tabular}{lrrrr}
\toprule
Integer digits & 1 & 2 & 3 & 4 \\
\midrule
TimeLens-100K target times & 29,526 & 140,488 & 23,158 & 0 \\
\midrule
Dropped digit, base answers & \multicolumn{4}{r}{273 of 489 (55.8\%)} \\
\bottomrule
\end{tabular}}
\end{table}

\paragraph{How frame times enter the model.}
TimeLens-8B and Qwen3-VL-8B receive 448 frames and their source times.
Qwen3-VL's video preprocessing groups frames in pairs and, before each pair's visual tokens, writes their mean time to one decimal place, for example \texttt{<1065.0 seconds>}.
A 448-frame video therefore carries 224 such timestamps. On long videos, TimeLens-8B therefore receives four-digit input times even though its training targets contain none.
InternVL3.5-8B receives a text timestamp for each frame; Qwen2.5-VL-7B encodes time in its position IDs (\cref{sec:sessions}).

\subsection{What the edits change in the answers}
\label{sec:what-changes}

\paragraph{The Qwen3-VL-8B edits change the digits after the separator.}
If the edits only changed the timestamp format, the edited answer would retain the base answer's numbers and replace the decimal point with a colon.
We compare the first two parsed timestamps after interpreting a decimal point as a colon, allowing a difference of at most one second in each time. For each edit, only 1 question whose separator changes meets this criterion (\cref{tab:qwen-split}). Most of the gain comes from answers that change the digits after the separator, sometimes changing the first number as well.
The separator-only test (\cref{sec:supp-separator}; main-paper Sec.~5.3) likewise improves generation, whereas reading the original numbers as minutes lowers accuracy.
``Minutes written as seconds'' names the diagnosed error; the repair also changes which digits follow the separator.

\begin{table*}[t]
\centering
\caption{\textbf{Qwen3-VL-8B gains grouped by changes in the answer.} Groups are assigned in row order. The first row allows a difference of at most one second in each of the first two parsed timestamps after interpreting a decimal point as a colon. ``First number'' means the digits before the separator in the first timestamp. Each group reports its question count, mean IoU change and share of the total gain (run 3: $+1.30$; run 4: $+2.88$). Shares are rounded, so they may not sum to 100\%. Evaluated on A30 GPUs.}
\label{tab:qwen-split}
\small
\setlength{\tabcolsep}{3pt}
\begin{tabular}{lrrrrrr}
\toprule
 & \multicolumn{3}{c}{Query edit (run 3)} & \multicolumn{3}{c}{Value/output edit (run 4)} \\
Answer change & $n$ & $\Delta$ & share & $n$ & $\Delta$ & share \\
\midrule
New separator, times within 1\,s & 1 & $-0.2$ & 0\% & 1 & $-0.2$ & 0\% \\
New separator, same first number, other changes & 83 & $+8.7$ & 46\% & 177 & $+7.7$ & 39\% \\
New separator, different first number & 63 & $+12.4$ & 49\% & 116 & $+10.4$ & 35\% \\
Same separator & 1,068 & $+0.1$ & 5\% & 921 & $+1.0$ & 27\% \\
\bottomrule
\end{tabular}
\end{table*}

\paragraph{Completing timestamps accounts for most of the gain of run 17.}
Run 17 improves IoU on 324 questions (27\%), leaves it unchanged on 778 (64\%), and worsens it on 113 (9\%); \cref{fig:qual-failures} shows the four largest losses. The improvements contribute $+9.33$ to mIoU and the losses subtract 1.44, leaving $+7.89$ overall.
Run 17 changes 592 of the 1,215 answers. Both timestamps gain digits in 134 answers (mean gain $+33.6$, accounting for 47\% of the total $+7.89$); one timestamp gains digits in 93 ($+26.6$, 26\%). The other 365 changed answers gain $+7.1$ $[+3.6, +11.1]$ on average, 27\% of the total gain.
Of the edited answers, 224 gain overlap with the true event and 24 lose it.
Run 16 changes 219 answers; completing timestamps accounts for 80\% of its $+3.13$ gain.

On videos of 10 to 30 minutes, the four largest run 17 gains all concern events covering nearly the whole video: the base answers lie in the first seconds, and the edit extends them across the video. As the largest gains, they are not typical.

Run 17 leaves 623 answers identical to the base answers. In four of them drawn at random, three videos last a few minutes, and in the fourth the event occurs in the first seconds; none needs a longer timestamp. \Cref{fig:qual-random} shows six test questions drawn at random, with both improvements and remaining errors.

\paragraph{Recall at IoU thresholds.}
\Cref{tab:recall} reports R1@0.3, 0.5 and 0.7 alongside mIoU. R1@0.3 is the share of questions whose answer reaches an IoU of at least 0.3, and likewise for 0.5 and 0.7.
For a question with several true spans, we compare their union with the predicted span, after rounding the predicted start down and its end up to whole seconds. The lower part of \cref{tab:recall} reports the 422 questions about videos of at least 30 minutes separately.

\paragraph{Bootstrap units.}
VUE-TR contains 246 test videos and VUE-TR-V2 contains 497, with 78 shared videos. The main paper resamples videos separately within each dataset: 743 units, including two units for each shared video. \Cref{tab:answer-controls} instead resamples the 665 distinct videos.
Using these 665 units widens run 17's interval over the base model from $[+6.28, +9.46]$ to $[+6.14, +9.80]$, and over the times-ten rule from $[+0.81, +3.81]$ to $[+0.72, +3.91]$. It changes no significance conclusion for runs 16 and 17 against either comparison.

\begin{table}[t]
\centering
\caption{\textbf{Recall at IoU thresholds.} Top: all 1,215 test questions. Bottom: the 422 questions about videos of at least 30 minutes. TimeLens-8B uses B200 GPUs and Qwen3-VL-8B uses A30 GPUs.}
\label{tab:recall}
\small
\setlength{\tabcolsep}{4pt}
\resizebox{\columnwidth}{!}{%
\begin{tabular}{llrrrr}
\toprule
Model & Weights & R1@0.3 & R1@0.5 & R1@0.7 & mIoU \\
\midrule
TimeLens-8B & base & 37.94 & 29.14 & 20.16 & 29.55 \\
 & run 16 edit & 42.14 & 32.35 & 22.39 & 32.68 \\
 & run 17 edit & 47.24 & 36.30 & 26.34 & 37.44 \\
 & LoRA & 49.63 & 38.85 & 27.90 & 38.94 \\
\midrule
Qwen3-VL-8B & base & 32.35 & 23.37 & 16.21 & 25.02 \\
 & query edit & 34.16 & 24.61 & 16.71 & 26.32 \\
 & value/output edit & 35.88 & 25.84 & 18.52 & 27.90 \\
 & LoRA & 48.56 & 37.53 & 27.41 & 37.99 \\
\midrule
\multicolumn{6}{l}{Videos of at least 30 minutes (TimeLens-8B)} \\
 & base & 8.29 & 5.69 & 3.79 & 6.79 \\
 & run 16 edit & 16.35 & 12.80 & 8.06 & 13.05 \\
 & run 17 edit & 31.52 & 23.46 & 16.82 & 25.06 \\
 & LoRA & 37.20 & 29.38 & 21.09 & 29.69 \\
\bottomrule
\end{tabular}}
\end{table}

\subsection{Where TimeLens-8B decides to end the number}
\label{sec:patching}

An edit can work at a layer other than the one that localisation methods point to~\cite{hase2023localization}. We test which blocks affect the decision to end a timestamp after three digits.
We identify 273 \emph{failure positions} where the base model writes a decimal point after three digits although the true time is at least 1,000 seconds and within 15\% of ten times the predicted integer part. We use 120 positions from 73 videos, at most three per video.
The 42 \emph{donor positions} follow the third digit of correctly written four-digit times.
Teacher-forcing the base answer, we measure the log-odds of a digit over a decimal point,
$\mathrm{LD} = \log \sum_{d} p(d) - \log p(\texttt{.})$ over the ten digits $d$; positive LD favours a digit.
We use mean activation patching: we replace one attention or MLP block's output at each failure with its mean at donor positions. The control instead uses its mean at failures from other videos.

\begin{figure}[!t]
\centering
\includegraphics[width=\linewidth]{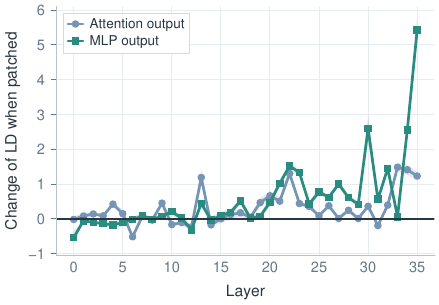}
\caption{\textbf{Late MLP blocks have the largest effect on the decimal-point decision.} Change in LD (\cref{sec:patching}) after replacing one layer's attention or MLP output at 120 failure positions with its mean at correct four-digit positions. Bands show 95\% bootstrap intervals over 73 videos. The largest effects occur in MLP layers 35, 30 and 34 (\cref{tab:patching}); runs 15 to 17 edit layer 34's MLP output projection. Evaluated on A30 GPUs.}
\label{fig:layer-patching}
\end{figure}

The base model favours a decimal point ($\mathrm{LD} = -8.22$ $[-8.85, -7.56]$). The largest patching effects occur in MLP layers 35, 30 and 34, the last of which runs 15 to 17 edit (\cref{tab:patching,fig:layer-patching}).
Layers 19 to 23, which overlap steering's window of layers 18 to 22 (\cref{tab:steering}), contribute at most $+1.52$ per block: $+7.80$ summed over their ten blocks, compared with $+17.36$ for layers 30 to 35. The first 13 layers each contribute less than $+0.5$; the control changes LD by less than 0.6 everywhere.
With any single block patched, the model continues the number at no more than 5.8\% of failure positions, compared with 63.3\% under run 17's edit.
This places the decision late in the network. However, patching favours blocks near the output because they affect logits directly; it does not identify where information about timestamp length is first lost.

\begin{table}[t]
\centering
\caption{\textbf{Replacing one block's output at the decimal-point decision.} Change in LD, the log-odds of continuing the number, at 120 failure positions when the block output is replaced by its mean at correct four-digit positions. Intervals are 95\% bootstrap intervals over 73 videos. A30 GPUs, bf16.}
\label{tab:patching}
\small
\setlength{\tabcolsep}{4pt}
\begin{tabular}{lrl}
\toprule
Patched block & $\Delta$LD & 95\% interval \\
\midrule
Layer 35 MLP & $+5.44$ & $[+5.09, +5.77]$ \\
Layer 30 MLP & $+2.60$ & $[+2.25, +2.97]$ \\
Layer 34 MLP (edited) & $+2.56$ & $[+2.32, +2.79]$ \\
Layer 22 MLP & $+1.52$ & $[+1.41, +1.63]$ \\
Layer 33 attention & $+1.49$ & $[+1.36, +1.61]$ \\
Layer 32 MLP & $+1.45$ & $[+1.35, +1.55]$ \\
Layer 34 attention & $+1.41$ & $[+1.26, +1.56]$ \\
Layers 0 to 12, each & $<+0.5$ & \\
\midrule
Run 17's edit & $\mathbf{+13.20}$ & $[+12.14, +14.29]$ \\
\bottomrule
\end{tabular}
\end{table}

\subsection{When the edit of run 17 changes the decimal-point decision}
\label{sec:selectivity}

The edits of runs 15 to 17 share a direction $u$ that lowers the decimal-point logit relative to digits. In the factorization below, $u$ uses the opposite sign convention to the construction direction $u$ in \cref{sec:sessions}. This shared direction alone does not validate the diagnosis: any session targeting that logit could find it.
Run 17's layer-34 update is effectively rank one, $\Delta W \approx \sigma\,u z^{\top}$ (99.99\% of its squared norm in the top singular pair). Its unit right singular vector $z$ selects when it acts. It lies in the MLP output projection's input space and equals the residual-space selector $v$ of \cref{sec:sessions} mapped through $W^{\top}$, up to sign and scale. For projection input $h$, the addition is $\sigma\,u\,(z^{\top} h)$.
Three controls retain $u$: a random selector $\sigma\,u\,r^{\top}h$ with random unit vector $r$; a mean selector $\sigma\,u\,m^{\top}h$ with unit mean input $m$; and a constant bias $c\,u$, with $c$ from 0.25 to 4 times the edit's mean addition at failure positions.
We test the 120 failure positions of \cref{sec:patching} and 120 positions with \emph{correct decimals}, where the model correctly ends the integer part of a number below 1,000 seconds.

With run 17's edit, the model writes another digit at 63.3\% of failure positions and only 2.5\% of positions where the decimal point is correct (\cref{tab:selectivity}). A constant bias that makes the model continue the number at least as often at failure positions also does so at 52.5\% of positions where the decimal point is correct. Weaker biases continue the number at no more than 9.2\% of failure positions.
The random selector has no effect. The mean selector has a large component of the opposite sign and makes every decision favour the decimal point.
Thus, the editor's selector distinguishes failure positions from positions where the decimal point is correct more effectively than these controls. This is consistent with its construction orthogonal to states at correct decimal points (\cref{sec:sessions}). These are single decisions under teacher forcing; \cref{sec:answer-controls} tests generation.

\begin{table}[t]
\centering
\caption{\textbf{Run 17 replaces most premature decimal points with digits and retains most correct decimal points.} Percentage of positions at which the model continues the number under each intervention, among 120 failure positions and 120 positions where the decimal point is correct. Positions come from TimeLens-8B's base test answers; the base rate is 0\% in both groups. Controls retain the edit's output direction. At $\times 4$, the model instead predicts other tokens and does not continue the number. A30 GPUs, bf16.}
\label{tab:selectivity}
\small
\setlength{\tabcolsep}{4pt}
\begin{tabular}{lrr}
\toprule
Change at layer 34 & Failures & Correct decimals \\
\midrule
Run 17's edit & 63.3\% & 2.5\% \\
Random selector & 0\% & 0\% \\
Mean selector & 0\% & 0\% \\
Constant bias $\times 0.25$ & 2.5\% & 0.8\% \\
Constant bias $\times 0.5$ & 9.2\% & 0.8\% \\
Constant bias $\times 1$ & 98.3\% & 52.5\% \\
Constant bias $\times 2$ & 100\% & 100\% \\
Constant bias $\times 4$ & 0\% & 0\% \\
\bottomrule
\end{tabular}
\end{table}

\subsection{Testing the selector controls during generation}
\label{sec:answer-controls}

We repeat the controls of \cref{sec:selectivity} during generation on all 1,215 test questions, using 448 frames, greedy decoding and at most 256 new tokens.
All conditions run on the same V100 GPUs in fp16. Run 17 gains $+7.34$ $[+5.69, +9.10]$ over the base model here, compared with $+7.89$ on B200 GPUs (main-paper Table~1).
The controls retain run 17's output direction $u$ at layer 34. Two add a constant bias $c\,u$ at every position, with $c = 0.5\,c_0$ or $c_0$; $c_0$ is the bias strength labelled $\times 1$ in \cref{tab:selectivity}. Here $c_0 = 201.69$ is the edit's mean addition $\sigma\,(z^{\top}h)$ at the 120 failures of \cref{sec:patching}, measured in that experiment and not chosen to maximise any score.
The third replaces the selector with a random unit vector. We omit the mean selector because it already favours the decimal point at every tested decision (\cref{tab:selectivity}).

The half-strength constant bias is the strongest control, gaining $+2.29$, about a third of run 17's gain (\cref{tab:answer-controls}). Run 17 exceeds it by $+5.05$ $[+3.66, +6.56]$ overall and $+14.71$ $[+11.46, +18.07]$ on videos of 30 minutes or more.
Below 10 minutes, neither differs significantly from the base model (run 17 $+0.62$, bias $+0.81$), or from each other ($-0.19$ $[-0.86, +0.49]$).
The full-strength constant bias changes every answer and degrades mIoU; the random selector also lowers it slightly.
The constant biases add digits in 82 of 226 changed answers at half strength and 721 of 1,215 at full strength, removing none. Run 17 adds digits in 240 of 575 changes; the random selector removes digits in 38 of 82 changes. The selector is therefore needed for the full gain during generation.

\begin{table*}[t]
\centering
\caption{\textbf{Run 17 outperforms controls with the same output direction during generation.} Test mIoU on 1,215 questions, gains over the base model, and run 17's gain over each control, with 95\% bootstrap intervals over the 665 distinct videos. The final column gives the latter gain on 422 questions about videos of 30 minutes or more, without intervals. Bold marks intervals that exclude zero. All conditions use the same V100 GPUs in fp16.}
\label{tab:answer-controls}
\small
\setlength{\tabcolsep}{2.5pt}
\begin{tabular}{lrrrr}
\toprule
Weights & mIoU & $-$ base & run 17 $-$ control & $\geq$30 min \\
\midrule
Base model & 29.59 & & $\mathbf{+7.34}$ {\scriptsize$[+5.69, +9.10]$} & $+17.53$ \\
Run 17's edit & 36.93 & $\mathbf{+7.34}$ {\scriptsize$[+5.69, +9.10]$} & & \\
Constant bias $0.5\,c_0\,u$ & 31.88 & $\mathbf{+2.29}$ {\scriptsize$[+1.37, +3.32]$} & $\mathbf{+5.05}$ {\scriptsize$[+3.66, +6.56]$} & $+14.71$ \\
Constant bias $c_0\,u$ & 10.21 & $\mathbf{-19.38}$ {\scriptsize$[-22.03, -16.83]$} & $\mathbf{+26.72}$ {\scriptsize$[+24.29, +29.26]$} & $+21.02$ \\
Random selector & 28.74 & $\mathbf{-0.85}$ {\scriptsize$[-1.37, -0.41]$} & $\mathbf{+8.19}$ {\scriptsize$[+6.43, +10.08]$} & $+18.47$ \\
\bottomrule
\end{tabular}
\end{table*}

\subsection{Short-video grounding and other video tasks}
\label{sec:locality}

\paragraph{TimeLens-Bench.}
We evaluate the base models and edits on the full TimeLens-Bench~\cite{zhang2026timelens}: Charades-TimeLens (3,363 queries), ActivityNet-TimeLens (4,500) and QVHighlights-TimeLens (1,541).
We follow the released code of the official TimeLens-Bench evaluation: 2 frames per second, at least 64 visual tokens per frame, at most 14,336 per video, the TimeLens grounding prompt, and greedy decoding. We score the first time pair in each answer after rounding to whole seconds.
Our base scores are within 0.5 mIoU of the published results. TimeLens-8B scores 55.23, 52.78 and 65.32 (published: 55.2, 53.2, 65.5); Qwen3-VL-8B-Instruct scores 48.34 and 59.38 on Charades- and QVHighlights-TimeLens (48.3, 59.4).

\Cref{tab:locality} reports the changes. Run 16 shows no significant difference on any benchmark. Run 17 shows none on two benchmarks and loses $0.49$ on ActivityNet-TimeLens.
The Qwen3-VL-8B query edit improves both evaluated benchmarks. The value/output edit improves QVHighlights-TimeLens and shows no significant change on Charades-TimeLens.
The latter result ($-0.06$ on all 3,363 questions under the official evaluation) differs from the main paper's $-1.01$. That result uses 297 questions chosen as opposite-action pairs and our own evaluation code (\cref{sec:headroom}).
The main paper's Qwen3-VL-8B LoRA improves both short-video benchmarks ($+4.31$, $+6.18$).

\begin{table*}[t]
\centering
\caption{\textbf{Changes in short-video grounding on TimeLens-Bench.} Columns are the TimeLens versions of each dataset. Each model's first row gives base mIoU; later rows give paired changes under the official evaluation, with 95\% bootstrap intervals over videos. Bold marks gains whose intervals lie above zero. A30 GPUs, bf16. Qwen3-VL-8B was not evaluated on ActivityNet-TimeLens.}
\label{tab:locality}
\small
\setlength{\tabcolsep}{2pt}
\begin{tabular}{llrrr}
\toprule
Model & Weights & Charades & ActivityNet & QVHighlights \\
\midrule
TimeLens-8B & base & 55.23 & 52.78 & 65.32 \\
 & run 17 & $-0.21$ {\scriptsize$[-0.51, +0.09]$} & $-0.49$ {\scriptsize$[-0.94, -0.04]$} & $+0.27$ {\scriptsize$[-0.34, +0.88]$} \\
 & run 16 & $+0.13$ {\scriptsize$[-0.02, +0.29]$} & $+0.18$ {\scriptsize$[-0.04, +0.41]$} & $+0.03$ {\scriptsize$[-0.33, +0.38]$} \\
\midrule
Qwen3-VL-8B & base & 48.34 & -- & 59.38 \\
 & query edit & $\mathbf{+0.60}$ {\scriptsize$[+0.33, +0.87]$} & -- & $\mathbf{+0.86}$ {\scriptsize$[+0.17, +1.56]$} \\
 & value/output & $-0.06$ {\scriptsize$[-0.25, +0.13]$} & -- & $\mathbf{+1.63}$ {\scriptsize$[+0.84, +2.47]$} \\
 & LoRA & $\mathbf{+4.31}$ {\scriptsize$[+3.52, +5.12]$} & -- & $\mathbf{+6.18}$ {\scriptsize$[+4.66, +7.67]$} \\
\bottomrule
\end{tabular}
\end{table*}

\paragraph{Video question answering.}
The main paper lists general video question answering as untested; we report it here.
We evaluate MVBench~\cite{li2024mvbench} with its official prompt, 16 frames and letter answers. TimeLens-8B scores 65.55\% on 3,800 questions across 19 of the 20 tasks. We cannot run Fine-grained Pose because its NTU RGB+D videos are not distributed with the benchmark.
On B200 GPUs in bf16, run 17 shows no significant change ($+0.11$ $[-0.21, +0.42]$, with 51 changed answers). Run 16 improves slightly ($+0.21$ $[+0.03, +0.42]$), as does the LoRA model of \cref{sec:lora-tl8} ($+0.76$ $[+0.29, +1.26]$).

\begin{figure}[t]
\centering
\includegraphics[width=\linewidth]{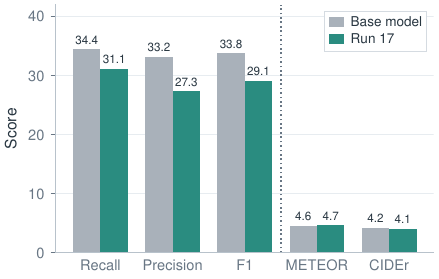}
\caption{\textbf{Run 17 worsens timestamp localisation without a significant change in caption quality.} ActivityNet Captions evaluation on 1,455 validation videos (\cref{sec:locality}). Left of the dotted line: localisation recall, precision and F1. Right: METEOR and CIDEr. Bars show scores over all videos, with F1 from the pooled recall and precision. The mean per-video F1 changes by $-3.63$ $[-4.84, -2.41]$; METEOR changes by $+0.12$ $[-0.10, +0.34]$.}
\label{fig:captioning}
\end{figure}

\begin{figure*}[!t]
\centering
\includegraphics[width=\textwidth]{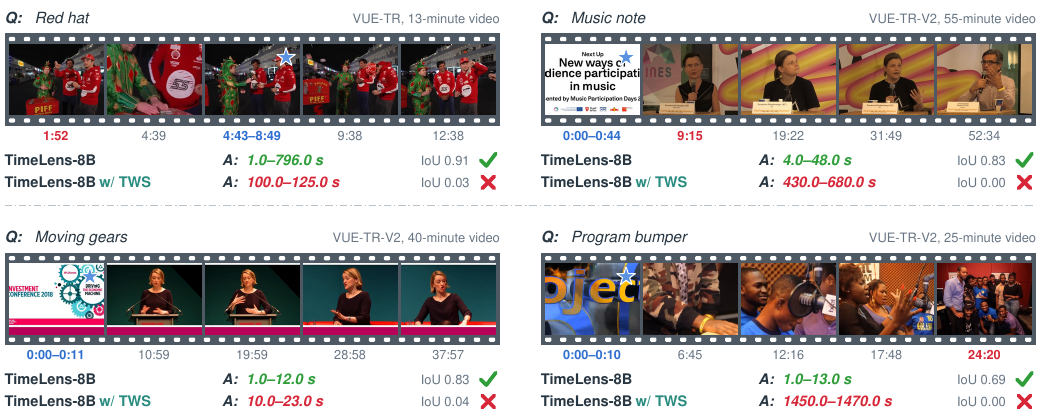}
\caption{\textbf{Failure cases.} The four largest run 17 IoU losses among 1,215 test questions, selected by rules fixed before viewing. Each strip shows five frames in time order. A star marks a frame in the true event (blue span); red timestamps mark incorrect answer centres. Answers as written and their IoUs (0 to 1) appear below; a check denotes IoU at least 0.5. In three cases, the base model finds a first-minute event but run 17 answers later, twice by extending a number (430.0 and 1450.0). In the first case, the base answer covers the whole video and most of ten true spans; run 17 returns a 25-second span.}
\label{fig:qual-failures}
\end{figure*}

\begin{figure*}[t]
\centering
\includegraphics[width=\textwidth]{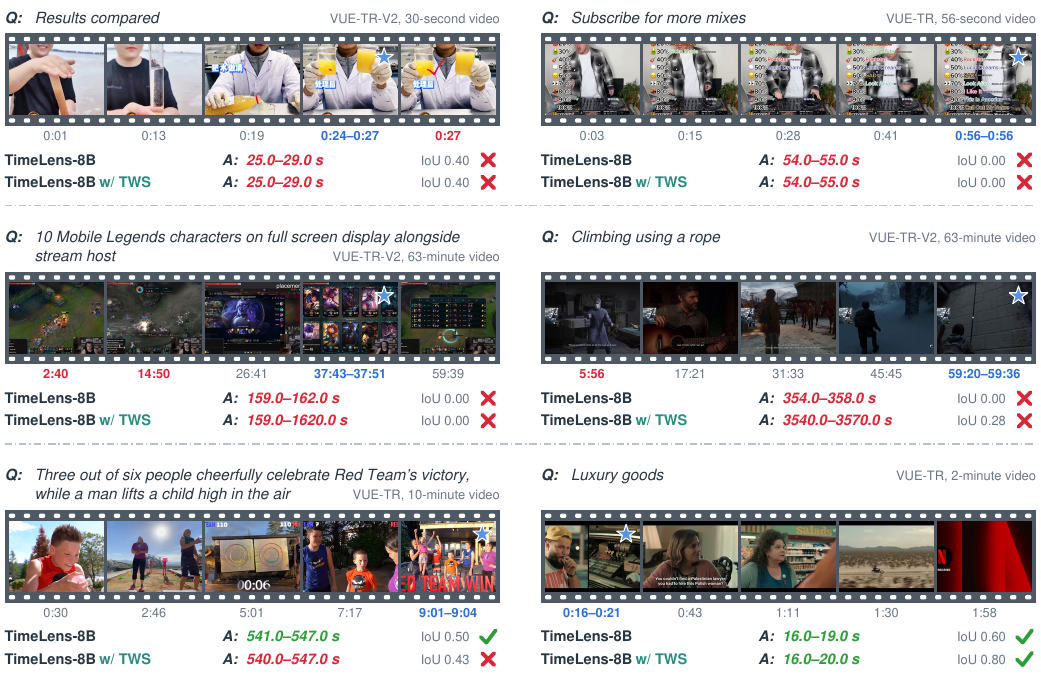}
\caption{\textbf{Six randomly sampled test questions.} The first six questions of a random draw from all 1,215 questions (seed 0), without further selection and with the rules fixed before viewing. Each strip shows five frames in time order. A star marks a frame in the true event (blue span); red timestamps mark incorrect answer centres. Answers as written and their IoUs (0 to 1) appear below; a check denotes IoU at least 0.5. Run 17 changes little on the four videos of up to ten minutes: one answer improves and one worsens slightly. On both hour-long videos it adds a digit, moving one answer onto the event (IoU 0 to 0.28) while leaving the other wrong.}
\label{fig:qual-random}
\end{figure*}

\paragraph{Timestamped captioning.}
For ActivityNet Captions~\cite{krishna2017dense}, we use TimeChat's dense-captioning instruction~\cite{ren2024timechat} to ask TimeLens-8B for events and their start and end times in seconds.
We evaluate the 1,455 ActivityNet-TimeLens validation videos with the official TimeLens frame sampling and greedy decoding capped at 512 new tokens. We run the official evaluator, updated to Python 3, with both reference sets and average over temporal IoU (tIoU) thresholds 0.3, 0.5, 0.7 and 0.9. Because this covers only 1,455 of 4,917 validation videos, the scores are not comparable with published results.
The base model scores 4.57 METEOR and 4.20 CIDEr, with localisation recall 34.43 and precision 33.16.
Run 17 shows no significant METEOR change (4.69; per-video gain $+0.12$ $[-0.10, +0.34]$), but localisation worsens (\cref{fig:captioning}): per-video F1 changes by $-3.63$ $[-4.84, -2.41]$, recall falls to 31.11 and precision to 27.32.
The loss comes from timestamps beyond the video end: 18.6\% of run 17's times, compared with 0.3\% of the base model's, across 391 videos rather than 30. Some answers continue listing events beyond the end. These 391 videos account for 96\% of the loss (mean F1 change $-13.04$); the other 1,064 change by $-0.18$.
Run 17 also starts 497 answers with the prompt's example times, ``90 - 102 seconds'', compared with 132 base answers.
These evaluations use B200 GPUs in bf16. Run 16 and LoRA were not evaluated on this task.

\subsection{LoRA on TimeLens-8B}
\label{sec:lora-tl8}

We train TimeLens-8B with the same LoRA training settings as the main paper's Qwen3-VL-8B experiment: rank 16, $\alpha = 32$, query, key, value and output projections in every language-model layer, learning rate $10^{-4}$, and two epochs (120 updates) on the 480 practice questions.
We evaluate all 1,215 test questions on the B200 machine, pairing with the base answers from run 17's evaluation. We also evaluate LoRA combined with run 17's edit, which acts on a tensor untouched by LoRA.

LoRA gains $+9.38$ $[+7.42, +11.39]$, exceeding run 17 by $+1.50$ $[+0.16, +2.86]$ overall and $+4.63$ on videos of 30 minutes or more (\cref{tab:lora-tl8}). Adding run 17's edit to the LoRA model changes mIoU by $-2.31$ $[-3.43, -1.20]$; the combination therefore performs worse overall.
For short-video grounding, the B200 base model scores 55.29, 52.98 and 65.64 on Charades-, ActivityNet- and QVHighlights-TimeLens. LoRA shows no significant change on Charades-TimeLens ($-0.23$ $[-0.83, +0.37]$). It changes ActivityNet-TimeLens by $-1.06$ $[-1.69, -0.45]$ and QVHighlights-TimeLens by $-2.02$ $[-3.01, -1.04]$; on A30 GPUs, neither edit changes QVHighlights-TimeLens significantly (\cref{tab:locality}). LoRA gains $0.76$ on MVBench.
Thus, gradient training gives a larger grounding gain on TimeLens-8B, as on Qwen3-VL-8B. Run 17 provides a diagnosis and a rank-1 change to one matrix, with smaller effects on short-video grounding but harm to timestamped captioning (\cref{sec:locality}). The Qwen3-VL-8B LoRA, by contrast, also improves short-video grounding.

\begin{table}[t]
\centering
\caption{\textbf{TimeLens-8B LoRA compared with run 17's edit.} Test mIoU on 1,215 questions, gain over the base model with 95\% bootstrap intervals over videos, and mIoU on the 422 questions about videos of 30 minutes or more. Bold marks the highest score in each column and gains whose intervals exclude zero. B200 GPUs, bf16.}
\label{tab:lora-tl8}
\small
\setlength{\tabcolsep}{3pt}
\resizebox{\columnwidth}{!}{%
\begin{tabular}{lrrr}
\toprule
Weights & mIoU & $-$ base & $\geq$30 min \\
\midrule
Base model & 29.55 & & 6.79 \\
Run 17's edit & 37.44 & $\mathbf{+7.89}$ {\scriptsize$[+6.28, +9.46]$} & 25.06 \\
LoRA & \textbf{38.94} & $\mathbf{+9.38}$ {\scriptsize$[+7.42, +11.39]$} & \textbf{29.69} \\
LoRA + run 17's edit & 36.62 & $\mathbf{+7.07}$ {\scriptsize$[+4.97, +9.23]$} & 27.85 \\
\bottomrule
\end{tabular}}
\end{table}

\subsection{Further sessions on TimeLens-8B}
\label{sec:new-sessions}

Runs 27 and 28 are two further GPT-6 Astra sessions on TimeLens-8B that test how reliably the repair is found. They use the settings of runs 15 to 17 (GPT-6 Astra through Codex, all 744 editable tensors, 50 forward-pass calls and the same 480 practice questions) and, like runs 16 and 17, B200 GPUs.
Each submitted edit is evaluated once on the 1,215 test questions and compared with a random rank-1 update of the same size (\cref{tab:new-sessions}).
Run 27 edits layer 34's MLP output, as runs 15 to 17 do; run 28 edits 48 columns of layer 32's MLP output. Both are rank-1 updates that lower the decimal-point logit relative to digit logits.
Both beat the base model and their random controls, so all five TimeLens-8B sessions with a tested edit do. Run 27 exceeds the times-ten rule ($+1.75$ $[+0.18, +3.29]$); run 28 performs worse ($-2.25$ $[-4.08, -0.49]$).
Runs 23 to 26 used a different interface with 10 forward-pass calls and ended without submitting an edit, so they have no test result.

\begin{table*}[t]
\centering
\caption{\textbf{Two additional TimeLens-8B sessions.} GPT-6 Astra through Codex, with 50 forward-pass calls and the settings of runs 15 to 17. Parentheses give relative update size. Bold marks intervals that exclude zero. Columns report the test gain over the base model and over a random rank-1 update of the same size, with 95\% bootstrap intervals over videos. B200 GPUs, bf16.}
\label{tab:new-sessions}
\small
\setlength{\tabcolsep}{2.5pt}
\begin{tabular}{rlrr}
\toprule
Run & Weight update & Test $\Delta$mIoU & Edit vs. control \\
\midrule
27 & rank-1 MLP output, layer 34 (0.06) & $\mathbf{+7.30}$ {\scriptsize$[+5.70, +8.98]$} & $\mathbf{+7.62}$ {\scriptsize$[+6.05, +9.26]$} \\
28 & rank-1 MLP output, layer 32 (0.05) & $\mathbf{+3.30}$ {\scriptsize$[+2.21, +4.50]$} & $\mathbf{+3.28}$ {\scriptsize$[+2.20, +4.46]$} \\
\bottomrule
\end{tabular}
\end{table*}

\subsection{Correction for multiple comparisons}
\label{sec:holm}

Of the main paper's 36 paired comparisons with 95\% intervals, 30 exclude zero.
We apply Holm's correction to all 36 at a family-wise level of 0.05, estimating two-sided $p$-values from the intervals by a normal approximation.
After correction, 27 remain significant, including every previously significant TimeLens-8B and Qwen3-VL-8B comparison. Three lose significance:
\begin{itemize}
\item run 18 versus its random control ($+2.06$; Holm $p = 0.085$);
\item Qwen3-VL-2B steering versus the base model ($+0.87$; Holm $p = 0.28$);
\item run 12 versus that steering ($+1.24$; Holm $p = 0.15$).
\end{itemize}
These three gains are not established after correction.
The long-video gain appears twice, from separate evaluations ($+18.51$ and $+18.27$ on videos of 30 minutes or more). Counting it once changes no significance conclusion.

\subsection{Scope of the claims}

\label{sec:scope}

The timestamp repairs in TimeLens-8B and Qwen3-VL-8B improve generated times but do not establish better event recognition (\cref{sec:what-changes,sec:headroom}). Run 18 addresses another failure, unfinished repeated reasoning (\cref{sec:sessions}).

\label{sec:clarifications}

\paragraph{Comparisons with simpler repairs.}
Run 17's overall advantages over steering and the times-ten rule survive Holm correction (\cref{sec:holm}). Its advantage over the rule comes from shorter videos; the long-video difference is not significant (\cref{sec:sessions}). The comparison with rules written by an editor from answers alone (\cref{sec:compute}) is outside that correction.
Weight edits need no extra decoding operation, but can affect other tasks (\cref{sec:locality}). Output rules affect only the answers to which they are applied.

\paragraph{Comparison with TimeLens2-8B.}
TimeLens2-8B's scores (\cref{sec:compute}) remain about 7 points above run 17's 42.7 and 33.5 on VUE-TR and VUE-TR-V2 (main-paper Table~2; other machines), and above TimeLens-8B LoRA (\cref{tab:lora-tl8}). The repairs do not close this gap. Under the TimeLens2 evaluation its scale ceiling is small (\cref{tab:protocol-census}), and in our setting runs 1 and 2 do not improve it.

\paragraph{Hardware in main-paper Table~1.}
Main-paper Table~1 takes the test gains of runs 3 and 4 from A30 GPUs ($+1.30$, $+2.88$) and their edit-versus-control gains from the A40 machine ($+1.95$, $+2.77$).
On A40, their gains over that machine's base model are $+1.91$ and $+2.85$, and their controls' gains $-0.04$ and $+0.08$.

\paragraph{Counting sessions.}
The 21 sessions in \cref{tab:sessions} are runs 1 to 19, 21 and 22. No session was given the number 20. Each submits at most one edit: 17 submitted an edit, and runs 5, 6, 11 and 22 did not.

\paragraph{Computing the query difference in Equation~1 of the main paper.}
We compute $\delta q_h$ by teacher-forcing the base model's own answers to run 3's 234 development questions.
At the first separator, the first ``\texttt{:}'' or ``\texttt{.}'' following a digit, we record head $h$'s query-projection output. This uses the layer's normalised input and precedes query normalisation and rotary embedding.
We subtract the mean output for genuine seconds answers from the mean for m:ss answers, using the editor's labels of each answer. Neither group includes collapsed answers with minutes written as seconds.
The probe $r_\ell$ is fitted on all positions of these same answers.

\paragraph{Question sets.}
\label{sec:question-sets}
The editor sees only the 480 labelled practice questions from 227 TimeLens2-93K videos of at least 10 minutes, none in the test sets. Each session splits these into development questions for building and choosing an edit, and held-out questions for checking it; \cref{tab:sessions} gives sizes and exceptions. The screening procedure uses 120 practice questions, 60 on videos of 10 to 30 minutes and 60 on longer videos, to select models and prompts (\cref{sec:sessions,sec:compute}). Each submitted edit is evaluated once on the 1,215 test questions, hidden from the editor.

\label{sec:further}

\paragraph{The separator result beyond the first minute.}
Of the 294 answers in main-paper Fig.~5a, 132 concern first-minute events. Their written number and true start minute are usually both 0. The 95 exact matches equal the random-pairing expectation within this group, so this group provides no independent evidence of timestamp knowledge.
For the other 162 answers, 36 first numbers match the true start minute (22.2\%). Chance rates are 2.4\% under random pairing within these answers, 5.1\% under pairing within video-length groups (0--10, 10--20, 20--30, 30--60 and over 60 minutes), and 4.8\% for a uniformly sampled minute of each video.
Within one minute, 58 match (35.8\%), compared with 7.0\% under random pairing and 13.5\% within video-length groups.
So much of the main paper's 45\% against 16\% comes from first-minute events; beyond the first minute, the exact-match rate is 4 to 9 times chance.

\paragraph{How many candidate edits a session evaluates.}
Runs 3 and 4 use 5 and 6 of their 10 forward-pass calls to evaluate edited weights, and TimeLens-8B runs 15, 16, 17, 27 and 28 use 30, 31, 38, 40 and 34 of their 50. The other sessions with call records use 3 to 7 of 10 (runs 1, 2 and 5), 23 and 24 of 30 (runs 18 and 19), 24 of 32 (run 22) and 32 to 35 of 50 (runs 7, 8, 9, 12, 14 and 21).
A call can score more than one edit, and an edit can be scored more than once, so these counts are not the number of distinct candidates; run 22, for example, scored 26 edits in its 24 such calls. Only the submitted edit is tested, once.
Session records keep turns and timings, not token counts, so \cref{tab:sessions} reports turns.

\paragraph{What the sessions diagnosed about the dropped digit.}
The training-range explanation in \cref{sec:why-digit} comes from our own count of training targets, not from the sessions. All five sessions diagnose an early decimal point. Run 15 also observes two-digit predictions for three-digit true times: the error ``is not simply a four-digit output limit''.
Increasing strength already causes side effects during editing: run 15 reports out-of-video spans and truncated practice answers; run 17 says it ``destroys timestamp termination'' (\cref{sec:locality}).

\paragraph{Model selection and the rule against gradients.}
The upper bounds used for screening rely on true answers (\cref{sec:sessions}), favouring cases where output rules may also work. The sessions test whether an editor can repair the error inside the model, not whether it beats those rules.
Our protocol, like the report's~\cite{takehi2026sparks}, prohibits backpropagation. It permits supervised fits such as ridge probes, and strength selection on practice questions. LoRA measures what gradient training achieves on the same questions (\cref{sec:lora-tl8}).

\paragraph{How video-language models write timestamps.}
TimeChat~\cite{ren2024timechat}, TimeLens-8B and Qwen3-VL write timestamps as numbers in text. Other approaches use absolute-time tokens (VTG-LLM~\cite{guo2025vtgllm}), frame numbers drawn on images (NumPro~\cite{wu2025numpro}), or a separate timestamp head (TRACE~\cite{guo2025trace}). VTimeLLM~\cite{huang2023vtimellm} and Momentor~\cite{qian2024momentor} train for moment boundaries and fine-grained temporal reasoning.
The errors repaired here arise only when timestamps are written as text. The separator experiment (main-paper Sec.~5.3) shows that a model can represent a time yet write it incorrectly. Du~\etal~\cite{du2026know} similarly find that a model can attend to an event yet write the wrong time.

{\small
\bibliographystyle{ieeenat_fullname}
\bibliography{main,supp}
}

\end{document}